\documentclass[sigconf]{acmart}
\AtBeginDocument{%
  \providecommand\BibTeX{{%
    \normalfont B\kern-0.5em{\scshape i\kern-0.25em b}\kern-0.8em\TeX}}}

\setcopyright{acmcopyright}
\copyrightyear{2023}
\acmYear{2023}
\acmDOI{XXXXXXX.XXXXXXX}

\acmConference[FAccT '23]{Conference on Fairness, Accountability and Transparency}{June 12--15,
  2023}{Chicago, IL, USA} 
\usepackage{graphicx}
\usepackage{booktabs}
\usepackage{array}
\newcolumntype{M}[1]{>{\centering\arraybackslash}p{#1}}
\usepackage{svg}
\usepackage{subfig}

\begin{document}


\title{On the Independence of Association Bias and Empirical Fairness in Language Models}

\author{Laura Cabello}
\email{lcp@di.ku.dk}
\affiliation{%
  \institution{University of Copenhagen}
  \city{Copenhagen}
  \country{Denmark}
}

\author{Anna Katrine Jørgensen}
\email{akj@di.ku.dk}
\affiliation{%
  \institution{University of Copenhagen}
  \city{Copenhagen}
  \country{Denmark}
}

\author{Anders Søgaard}
\email{soegaard@di.ku.dk}
\affiliation{%
  \institution{University of Copenhagen}
  \city{Copenhagen}
  \country{Denmark}
}


\begin{abstract}
The societal impact of pre-trained language models has prompted researchers to probe them for strong associations between protected attributes and value-loaded terms, from slur to prestigious job titles. Such work is said to probe models for {\em bias or fairness}---or such probes `into representational biases' are said to be `motivated by fairness'---suggesting an intimate connection between bias and fairness. We provide conceptual clarity by distinguishing between association biases \cite{10.1145/3514094.3534162} and empirical fairness \cite{shen-etal-2022-representational} and show the two can be independent. Our main contribution, however, is showing why this should {\em not} come as a surprise. To this end, we first provide a thought experiment, showing how association bias and empirical fairness can be completely orthogonal.
Next, we provide empirical evidence that there is no correlation between bias metrics and fairness metrics across the most widely used language models.
Finally, we survey the sociological and psychological literature and show how this literature provides ample support for expecting these metrics to be uncorrelated. 

\end{abstract}





\maketitle

\section{Introduction}\label{intro}

The prevalence of unintended social biases in pre-trained language models (PLMs) is alarming, since they impact millions, if not billions of people every day. In recent years, more and more NLP researchers have studied such biases, making up an estimated 6.3\% of the literature in 2022 \cite{ruder-etal-2022-square}. 
Much of this work has focused on what \citet{crawford2017trouble} called {\em representational bias}, 
which manifests when portrayals of certain demographic groups are discriminatory. In NLP, representational bias often arises when associations between a protected attribute, e.g., gender, and certain concepts, e.g., job titles, are captured in the model space. Thus, to avoid ambiguity, we will refer to this type of bias as {\em association bias}, following \citet{chaloner-maldonado-2019-measuring}.

Association bias is often confused with what is sometimes referred to as performance disparity \cite{DBLP:conf/icml/HashimotoSNL18} or {\em empirical fairness} \cite{shen-etal-2022-representational}, i.e., performance differences across end user demographics. 
Or mitigating association bias is assumed to improve empirical fairness \cite{chen-etal-2020-analyzing,friedrich-etal-2021-debie,cao-etal-2022-intrinsic,dayanik-pado-2020-masking,Castelnovo_2022,Benchmarking2021_2723d092}. 
Note that most fairness metric focus on some form of equal performance and differ only in whether they focus on precision, recall or balancing the two \cite{barocas-hardt-narayanan}. Empirical fairness refers to equal performance as measured by {\em de facto} standard metrics and is arguably the most common fairness metric \cite{pmlr-v97-williamson19a,barocas-hardt-narayanan}.

In this paper, we will show that the two phenomena, association bias and empirical fairness, are often completely independent matters.\footnote{Note that the distinction between association bias and empirical fairness---between how expressions referring to demographic groups are encoded, and how these groups are treated as end users---is different from another distinction made in recent work \cite{biased-rulers, goldfarb-tarrant-etal-2021-intrinsic, kaneko-etal-2022-debiasing} between intrinsic and extrinsic bias: Intrinsic bias, here, is what we call representational bias, whereas extrinsic bias refers to performance differences on sentences containing entities referring to different demographic groups.} We devise a thought experiment (\S\ref{sec:theory}) to illustrate this, but also present a series of experiments (\S\ref{sec:practice}) to show that results obtained the way association bias is normally measured, do not correlate with results obtained the way empirical fairness is normally measured. 

Our main contribution, however, is to show that this should not come as a surprise. 
Research on mitigating association bias and empirical fairness is often motivated by fairness concerns, and bias and fairness are often considered near-synonymous terms in the research literature: 
Researchers have, for example, said that bias {\em causes}~unfairness \cite{chang-etal-2019-bias,friedrich-etal-2021-debie,Castelnovo_2022}. If this was the case, the independence of association bias and empirical fairness should come as a great surprise. However, the assumption that bias causes unfairness, is unwarranted, as we will see below, from a survey of relevant literature from the social sciences (\S\ref{sec:humans}). 
A causal link between association bias and empirical fairness would seem to require some sort of in-group affinity, i.e., that groups use terms relating to their in-group peers more and in different ways than outsiders, like, for instance, Democrats on Twitter mention Trump and the Republican party more often than their Republican counterparts \cite{Van_Duijnhoven2018}. This assumption, which we call the In-Group Affinity Assumption, seems intuitive, but without much support from the social sciences (\S\ref{sec:humans}).

\begin{figure}
    \centering
    \includegraphics[width=3in]{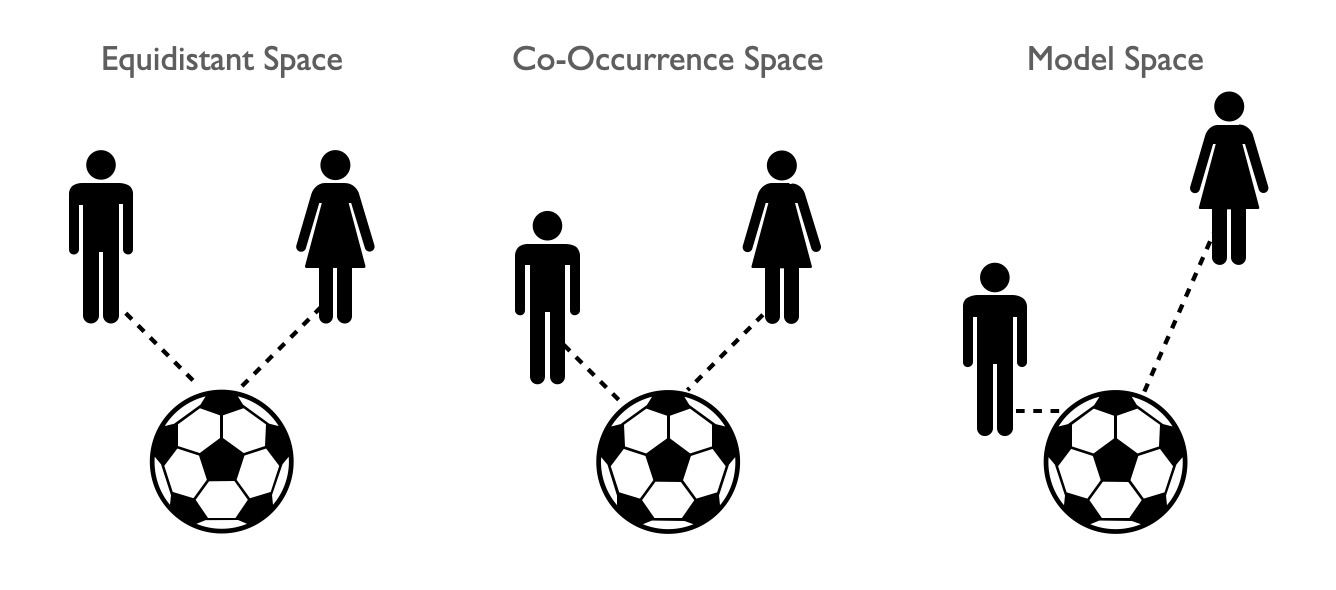}
    \caption{
    Association bias of group-related terms (e.g., {\em woman} and {\em man}) can be quantified as degree of isomorphism relative to an empirical ({\bf co-occurrence}) space or a normative, {\bf equidistant} space. The graph illustrates how {\em man} may be more strongly associated with {\em soccer} in a {\bf model}, less so empirically (the underlying data or real-world statistics), and not at all in an ideal world.}
    \label{fig:into}
    \Description{Association bias of group-related terms (e.g., woman and man) can be quantified as degree of isomorphism relative to an empirical (i.e.  co-occurrence) space or a normative, equidistant space. The graph illustrates how man may be more strongly associated with soccer in a model, less so empirically, and not at all in an ideal world.}
\end{figure}

\paragraph{Contributions} In \S\ref{sec2}, we define association bias and empirical fairness and discuss related work. When we talk about association bias, we refer to systematic biases in how words and phrases referring to demographic groups are encoded. Figure~\ref{fig:into} visualizes how models may exhibit biased associations because of sample biases, and may even amplify these. 
We define empirical fairness as equal performance across groups, because this is the most balanced and most widely applicable measure of fairness in NLP, except for specialized applications where equal base rates and calibration take priority over performance. We then move to study how association bias and empirical fairness relate. 
In \S\ref{sec:theory}, we show that theoretically, association bias and empirical fairness are completely independent. That is, mitigating association bias can hurt empirical fairness, and ensuring empirical fairness can introduce more bias.
\S\ref{sec:practice} shows there is no obvious correlation between results obtained from standard association bias measurements and results obtained from standard empirical fairness measurements of language models. 
Finally, \S\ref{sec:humans} surveys the social science literature for explanations on why association bias and empirical fairness may be less related (or related in less obvious ways) than multiple works in the NLP literature have assumed up to this point. The finding that association bias and empirical fairness are independent in this three-way investigation, should help push research horizons and provide strong motivation for targeting empirical fairness directly, as well as for seeing association bias mitigation, not necessarily as a way of promoting fairness, but rather as a way of preventing poor inferences and generation of stereotypical text.


\section{Definitions and Related Work}
\label{sec2}

In the NLP literature, bias and fairness are often conflated, or it is argued that one follows from the other, e.g., that we can ensure fairness by mitigating bias
\cite{chen-etal-2020-analyzing,friedrich-etal-2021-debie,cao-etal-2022-intrinsic,dayanik-pado-2020-masking,Castelnovo_2022,Benchmarking2021_2723d092}.
In contrast, we will show that this is not always the case, and (association) bias and (empirical) fairness often are independent or at odds.

\paragraph{Bias} Mitigating social biases in NLP models has become an important research goal \cite{shah-etal-2020-predictive, hutchinson-etal-2020-social, romanov-what-name}, but there is little consensus on how to evaluate such biases \cite{blodgett2020, karolina-survey}. We focus on association bias 
and show how, contrary to what seems to be popular belief, 
it is not unequivocally related to fairness; in fact, it is very often completely independent thereof.  

Association bias in a model refers to systematic differences in how words and phrases referring to demographic groups are encoded. Classical tests thereof include comparing the (cosine) distance of terms relating to protected attributes, e.g., {\em woman}~and {\em man}, or their vectors $\mathbf{v}_w$ and $\mathbf{v}_m$, to terms of particular interest, e.g., slur \cite{sap-etal-2019-risk}, sentiment \cite{DBLP:conf/icml/AliSEMMW22}, or job titles such as {\em doctor} ($\mathbf{w}_d$) \cite{zhao-etal-2018-learning}. Early papers would quantify bias with respect to, say, gender, as cosine similarities 
$(\mathbf{cos}(\mathbf{v}_m,\mathbf{w}_d)-\mathbf{cos}(\mathbf{v}_w,\mathbf{w}_d))$ 
\cite{caliskan-weat, BHATIA201746, zhao-etal-2018-learning, pmlr-v97-brunet19a, gonen-goldberg-2019-lipstick}, and by seeing whether the nearest neighbor of $\mathbf{w}_d+\mathbf{v}_w-\mathbf{v}_m$ would be {\em nurse} or another job stereotypically associated with women \cite{bolukbasi2016man}. In practice, NLP researchers have used tests such as the ones above for quantifying association bias \cite{caliskan-weat, BHATIA201746, zhao-etal-2018-learning, pmlr-v97-brunet19a, gonen-goldberg-2019-lipstick}. We will argue that such quantities are theoretically and, often practically, orthogonal to empirical fairness,
which we define in terms of differences in performance estimates across demographics, i.e., social groups \cite{pmlr-v97-williamson19a,barocas-hardt-narayanan,shen-etal-2022-representational}, often defined by the cross-product of a subset of protected attributes such as gender, age, or race.

\paragraph{Fairness} 
Fairness metrics come in multiple flavors, but are often divided in three: calibration-based, precision-based, and recall-based metrics. \citet{Miconi2017TheIO}, \citet{impossibility-friedler} and \citet{kleinberg-trdeoffs} show how pairs of fairness metrics can be mathematically incompatible, i.e., one type of fairness can rule out another. In fact, incompatibility holds for all pairs of metrics such that the two metrics are of different flavor, e.g., calibration-based and recall-based, unless the true base rates are identical across groups, or the classifier has perfect performance. Since the vast majority of NLP applications provide repetitive services, the quality of which can be measured against a gold standard, precision- and recall-based metrics are predominantly used in NLP. We follow several authors \cite{DBLP:conf/icml/HashimotoSNL18,hansen-sogaard-2021-lottery,chalkidis-etal-2022-fairlex} in using min-max differences in (the standard) performance (metric) as our go-to fairness metric.  
Relying on min-max difference captures the widely shared intuition that fairness is always in the service of the worst off group \cite{rawls_theory_1971}. For a discussion of available fairness metrics, and in what contexts they are relevant, see \citet{mehrabi-survey} and \citet{barocas-hardt-narayanan}. For a comparison of existing metrics used to quantify social biases in NLP, see \citet{10.1162/tacl_a_00425}.

\paragraph{Related Work} \citet{subpopulation-shift-fairness} study the effect of subpopulation shifts on performance disparities and show that these do not always relate in obvious ways.
\citet{goldfarb-tarrant-etal-2021-intrinsic} study the correlation between what they refer to as `intrinsic and extrinsic measures of representational bias'. Their intrinsic measures of representational bias amount to word association bias, but their extrinsic measures of representational bias are not empirical fairness measures. To see this, consider the coreference task used in \citet{goldfarb-tarrant-etal-2021-intrinsic}. \citet{goldfarb-tarrant-etal-2021-intrinsic} correlate intrinsic gender bias measures (cosine distances in static word embedding spaces) with coreference performance on sentences with female and male referents. We argue that in this case, empirical fairness would be performance on sentences {\em written by} female and male authors.\footnote{Or, alternatively, sentences {\em read by} female and male authors. The latter is rarely studied, as it requires reader statistics, e.g., from online media services.} \citet{goldfarb-tarrant-etal-2021-intrinsic} establish that there is no correlation between their two measures of representational bias. Their result superficially looks similar to and in agreement with ours, but is, in fact, unrelated. If anything, it shows that association bias has been assumed to correlate with many measures that it does not, in fact, correlate with. 
\citet{cao-etal-2022-intrinsic} and \citet{kaneko-etal-2022-debiasing} studied the same problem as \citet{goldfarb-tarrant-etal-2021-intrinsic}, but used contextualized token embeddings from PLMs rather than static word embeddings. They both found weak correlations between intrinsic and extrinsic evaluation measures. Again, we emphasize that these results do not contradict ours. 

\citet{shah-etal-2020-predictive} carefully avoid to discuss fairness, saying the fairness literature is outside the scope of their paper, but place outcome disparity (performance disparity) as a central motivation for social bias mitigation. They list four potential causes of outcome disparity: label bias, selection bias, bias amplification, and semantic (representation) bias. We show association (representation) bias and outcome disparity are theoretically, and also often practically, independent, questioning their fourth hypothesis. Moreover, we observe that outcome disparity can arise in the absence of {\em all}~of the above four factors. Say a group exhibits more variance than others, e.g., because of spelling variation in dyslexics. Even if dyslexics are represented proportionally or equally, they may still see worse performance with dyslexics than for non-dyslexics. 

Finally, \citet{shen-etal-2022-representational} show how a different form of representational fairness, i.e., whether protected author attributes can be detected from model representations, is also uncorrelated with empirical fairness. Together, our work and previous work \cite{goldfarb-tarrant-etal-2021-intrinsic,cao-etal-2022-intrinsic,kaneko-etal-2022-debiasing,shen-etal-2022-representational} establish that four common bias-related measures -- (i) association bias, (ii) performance on sentences with protected attribute terms (\citet{goldfarb-tarrant-etal-2021-intrinsic}'s extrinsic measure), (iii) decodability of protected attributes from representations, and (iv) empirical fairness are largely uncorrelated. Specifically, (i) is independent of (ii) and (iv), and (iii) is independent of (iv). 

Our work is motivated by the large-spread assumption that association bias and empirical fairness are causally related 
\cite{chen-etal-2020-analyzing,friedrich-etal-2021-debie,cao-etal-2022-intrinsic,dayanik-pado-2020-masking,Castelnovo_2022,liu-etal-2020-gender,qian2022perturbation,sun-etal-2019-mitigating,ross-etal-2021-measuring,bartl-etal-2020-unmasking}. 
\citet{bartl-etal-2020-unmasking}, for example, aspire ``to promoting fairness in NLP by exploring methods to measure and mitigate gender bias.'' \citet{ross-etal-2021-measuring} say they ``believe that by revealing biases, by providing tests for biases that are as focused as possible on the smallest units of systems, we can both assist the development of better models and allow the auditing of models to ascertain their fairness.'' 
\citet{sun-etal-2019-mitigating}, argue that ``biased predictions may discourage minorities from using those systems and having their data collected, thus worsening the disparity in the data sets'', equating biased predictions with unfair predictions.

All three sets of authors see bias as the primary cause of fairness. 
Showing such causation is not a given, and that in fact, association bias and empirical fairness need not even correlate and are often orthogonal, is an important correction to this literature, with potential consequences for research methodology, applications of NLP in the social sciences, as well as AI ethics and regulation. 
\vspace{-2mm}

\section{Association Bias and Empirical Fairness are Independent (in Theory)}
\label{sec:theory}

In this section, we produce a thought experiment---a synthetic model---to illustrate how bias and fairness can in fact be completely independent of one another. 
We construct a synthetic ternary (positive/negative/neutral) sentiment analysis model with a small feature space, including words that refer to demographic subgroups of a population. These words, denoting various groups, will be biased and associated with sentiment, because of biases in our training data. This assumption is also made in \citet{https://doi.org/10.48550/arxiv.2202.07304}, for example. These associations lead to biased likelihood estimates and would, in the context of a linear model, lead to differences in 
the degree of isomorphism relative to 
the group-specific subgraphs. We will show, however, that the resulting biases are independent of the group fairness of the model, i.e., to the min-max performance disparities across the same groups. 
Such a connection, if it exists, could be explained by an {\em in-group affinity}, which relies on the assumption that those biased terms are used by the in-group more frequently or in other ways than by other groups.

\smallskip
Say a population consists of members of groups $g_1,\ldots,g_4$, e.g., defined according to their address as {\em north}, {\em east}, {\em west} and {\em south}. Everyone speaks the same language and expresses sentiment with a vocabulary of seven words: $w_{g_1},\ldots,w_{g_4},w_5,w_6,w_7$. Except $w_6$ (positive) and $w_7$ (neutral), all words express negative sentiment, including the words that refer to (or are associated with) other demographic subgroups ($w_{g_i}$), for instance, {\em northern}, {\em eastern}, {\em western} and {\em southern}. The subgroups use the terms with the following probabilities (Table~\ref{tab:s41}): 


\begin{table}[h!]
\centering
\begin{tabular}{c|ccccccc}
     &$w_{g_1}$&$w_{g_2}$&$w_{g_3}$&$w_{g_4}$&$w_5$&$w_6$&$w_7$\\
     \hline $g_1$& 0.0&0.25&0.0&0.0&0.25&0.25&0.25\\          
     $g_2$& 0.0&0.0&0.25&0.0&0.25&0.25&0.25\\
     $g_3$& 0.0&0.0&0.0&0.25&0.25&0.25&0.25\\
     $g_4$& 0.25&0.0&0.0&0.0&0.25&0.25&0.25\\
\end{tabular}
\caption{Probability of a group $g_i$ using the word $w_j$ for expressing sentiment. Only $w_6$ (positive) and $w_7$ (neutral) express a non-negative sentiment.}
\label{tab:s41}
\end{table}

This data exhibits four representational biases, e.g., the association of $g_1$ with negative sentiment, the association of $g_2$ with negative sentiment, and so forth. If we have sufficient data, a simple model, e.g., a Naive Bayes classifier trained on simple 
bag-of-words representations, should induce the maximum likelihood estimates (where `0' denotes negative, `1' positive and `2' neutral sentiment) showcased in Table~\ref{tab:s42}.


\begin{table}[h!]
\resizebox{\columnwidth}{!}{
\centering
\begin{tabular}{c|ccccccc}    &$P(w_{g_1}|0)$&$P(w_{g_2}|0)$&$P(w_{g_3}|0)$&$P(w_{g_4}|0)$&$P(w_5|0)$&$P(w_6|0)$&$P(w_7|0)$ \\
     \hline $g_1$& 0.0&0.25&0.0&0.0&0.25&0.0&0.0\\  
     $g_2$& 0.0&0.0&0.25&0.0&0.25&0.0&0.0\\
     $g_3$& 0.0&0.0&0.0&0.25&0.25&0.0&0.0\\
     $g_4$& 0.25&0.0&0.0&0.0&0.25&0.0&0.0\\
\end{tabular}
}

\bigskip
\resizebox{\columnwidth}{!}{
\centering
\begin{tabular}{c|ccccccc}
&$P(w_{g_1}|1)$&$P(w_{g_2}|1)$&$P(w_{g_3}|1$)&$P(w_{g_4}|1)$&$P(w_5|1)$&$P(w_6|1)$&$P(w_7|1)$\\
     \hline $g_1$& 0.0&0.0&0.0&0.0&0.0&0.25&0.0\\          
     $g_2$& 0.0&0.0&0.0&0.0&0.0&0.25&0.0\\
     $g_3$& 0.0&0.0&0.0&0.0&0.0&0.25&0.0\\
     $g_4$& 0.0&0.0&0.0&0.0&0.0&0.25&0.0\\
\end{tabular}
}

\bigskip
\resizebox{\columnwidth}{!}{
\centering
\begin{tabular}{c|ccccccc}
&$P(w_{g_1}|2)$&$P(w_{g_2}|2)$&$P(w_{g_3}|2$)&$P(w_{g_4}|2)$&$P(w_5|2)$&$P(w_6|2)$&$P(w_7|2)$\\
     \hline $g_1$& 0.0&0.0&0.0&0.0&0.0&0.0&0.25\\          
     $g_2$& 0.0&0.0&0.0&0.0&0.0&0.0&0.25\\
     $g_3$& 0.0&0.0&0.0&0.0&0.0&0.0&0.25\\
     $g_4$& 0.0&0.0&0.0&0.0&0.0&0.0&0.25\\
\end{tabular}
}
\caption{Maximum likelihood estimates from a linear classifier on our synthetic data modelled in Table~\ref{tab:s41}.}
\label{tab:s42}
\end{table}

Now, say we employ an existing debiasing approach and manage to debias the model with respect to its representation of group $g_1$ by setting 
$P(w_{g_1}|0)=P(w_{g_1}|1)=P(w_{g_1}|2)$, which, in this case, would equal zero.
This would hurt performance on data from $g_4$ (bottom row), increasing the empirical risk on this sub-population, but more surprisingly, note that it would not help us on classifying the data from $g_1$. That is, an attempt to make the model fairer towards {\em north} by equalizing the use of the term {\em northern}, would result in increased unfairness towards members from {\em south}, who tend to use {\em northern} more often (and in a negative context).
Removing bias in how terms referring to a group are represented, only improves performance on data from members from that group, if these members use such in-group terms in non-standard ways, i.e., differently from everyone else. In the absence of this assumption, association bias and empirical fairness are orthogonal. We will refer to this assumption as the {\bf In-Group Affinity Assumption}. 

Note that while we make use of a linear model and likelihood estimates in our thought experiment, it would be very easy to translate this into a deep neural network and cosine distances instead. To see this, consider, for example, how any Naive Bayes model can be translated into a deep neural network, and how the differences in likelihood can, under such a translation, be translated into differences in cosine instances. 



\section{Association Bias and Empirical Fairness Scores are Uncorrelated (in Practice)}
\label{sec:practice}

In this section, we study whether association bias and empirical fairness are correlated in practice,i.e., when {\em actual} models are evaluated on {\em actual} data designed to probe bias and fairness.
We apply well-established metrics for measuring the two.
While bias and fairness can be studied with respect to any prospective attribute, the vast majority of NLP research has focused on (binary) gender \cite{sun-etal-2019-mitigating, karolina-survey}. Binary gender is often correlated with terms referring to occupations, e.g., the co-occurrence of {\em woman} and {\em man}--or {\em she}~and {\em he}--in the context of {\em nurse} and {\em doctor}. For convenience, we rely on existing benchmarks and do the same. It is important to remember, however, that bias and fairness may arise across any groups in society, and that all those defined in terms of protected attributes, e.g., race, religion, sexuality, or impairment, are legally irrelevant. As mentioned, the two--association bias and empirical fairness--are often conflated, or one is said to cause the other. This reflects an In-Group Affinity Assumption, saying that members of social groups refer to themselves more often or in different ways than other members of a linguistic community. If this were the case, mitigating biases would contribute positively to equal performance across groups. 

The analysis of these experiments concludes our three-way investigation of the In-Group Affinity Assumption and the independence of bias and fairness. All three perspectives suggest that NLP research should not further assume an intimate connection between the two.

\paragraph{Bias} To measure representational bias, we use three popular metrics, i.e., the Log Probability Bias Score (LPBS) proposed by \citet{kurita-etal-2019-measuring}, as well as two variants of the Word Embedding Association Test (WEAT) \cite{caliskan-weat} for assessing bias in contextual word representations:
the adaption proposed by \citet{tan-and-celis} (henceforth, WEAT$_T$), and the alternative suggested by \citet{lauscher-etal-2021-sustainable-modular} (henceforth, WEAT$_L$).
All these metrics rely on association tests to compute the relationship between a set of related targets \{$t_1, t_2, \ldots$\}, e.g., gender words, and attributes \{$a_1, a_2, \ldots$\}, e.g., occupation words, through definitions of template sentences designed to convey no meaning beyond that of the terms inserted into them.

\citet{kurita-etal-2019-measuring} use template sentences like 
$T$=``[TARGET] is a [ATTRIBUTE]''.
The target word is masked, and the attribute word is a place-holder for a specific word denoting an occupation, e.g., $T_m$=``[MASK] is a \emph{chef}''. LPBS uses the prior probability of the target word ($p_{prior}$), i.e., the probability of a target $t_i$ being generated when both $t_i$ and the attribute $a_j$ are masked, as a normalizer, and computes the association as the relative increase in log probability: 

\begin{equation}
    a^{lpbs}_{t_i, a_j} = \log\frac{p([\mbox{MASK}]=t_i|T_m)}{p_{prior}} 
\end{equation}

The difference between the relative increased log probability scores for two targets is the LPBS measure of bias. For linear models, this correlates strongly with the $\epsilon$-isometry of the target word subgraph relative to an equidistant space, if we make the centroid of the set of attribute vectors the reference point. For a non-linear language model, we can compute the $\epsilon$-isometry of its linear approximation. Table~\ref{tab:bias-metrics} are for the targets ``he'' and ``she''. 
A t-test is used to evaluate the statistical significance of the metric, in which the means of $a^{lpbs}_{he, a_j}$ and $a^{lpbs}_{she, a_j}$ are compared. We draw $10^5$ random permutations, meaning that the {\em p}-values observed will not be less than $10^{-5}$. 

\citet{tan-and-celis} follow the methodology of \citet{may-etal-2019-measuring}, who extended the WEAT metric to sentences (SEAT) inserting the word of interest in context templates such as 
$T$=``This is \_''. 
\citet{tan-and-celis} use the contextual embedding of the token of interest, instead of using the sentence encoding, to compute the cosine similarities (associations). 
\citet{lauscher-etal-2021-sustainable-modular} follow \citet{vulic-multisimlex} and average the pooled embeddings of the first four attention layers for the word of interest ($t_i$ or $a_j$) in a template without context, e.g, ``$[\mbox{CLS}]\ t_i\ [\mbox{SEP}]$''.
Both approaches report the \emph{effect size} \cite{caliskan-weat}, a normalized measure of how separated the association distributions of target and attributes are. The statistical significance of the associations is also computed with a permutation test as in \cite{caliskan-weat}. Both approaches are an instance of computing the $\epsilon$-isometry of the template sentence subgraphs in the cosine metric space. See Table~\ref{tab:bias-metrics} for empirical results.\footnote{PLM names follow the same nomenclature as in the Hugging Face Transformers library. The pre-trained models can be downloaded at \href{https://huggingface.co/models}{\nolinkurl{huggingface.co/models}}.} We see that results are somewhat mixed, with LPBS and the two variants of WEAT often disagreeing which models are more biased. All the metrics are evaluated on the same list of sixty attributes {--equally split into female and male stereotyped professions from the US bureau of labour--, provided in \cite{biased-rulers}.}


\begin{table}[htbp!]
\centering
\resizebox{\columnwidth}{!}{
\begin{tabular}{m{0.45\linewidth}M{0.21\linewidth}M{0.21\linewidth}M{0.21\linewidth}}
& $LPBS$ &$WEAT_L$ & $WEAT_T$ \\ \hline
bert-base-uncased& 0.86$^{*\phantom{*}}$& 1.01$^{*\phantom{*}}$& 0.33$\phantom{^{**}}$\\
bert-base-cased& 0.90$^{*\phantom{*}}$& 1.00$^{*\phantom{*}}$& 0.52$\phantom{^{**}}$\\
bert-large-uncased& 0.20$\phantom{^{**}}$& 0.83$^{*\phantom{*}}$& 0.73$^{*\phantom{*}}$\\
bert-large-cased& -1.10$^{*\phantom{*}}$& 0.60$\phantom{^{**}}$& 0.83$^{*\phantom{*}}$\\
bert-base-multilingual-cased& -1.98$^{*\phantom{*}}$& 0.36$\phantom{^{**}}$& 0.12$\phantom{^{**}}$\\\hline
distilbert-base-uncased& -0.46$^{*\phantom{*}}$& 0.79$^{*\phantom{*}}$& 0.58$\phantom{^{**}}$\\
albert-base-v2& -7.02$^{*\phantom{*}}$& 0.72$^{*\phantom{*}}$& 0.56$\phantom{^{**}}$\\
albert-large-v2& -1.58$^{*\phantom{*}}$& 0.84$^{*\phantom{*}}$& 0.61$^{*\phantom{*}}$\\
albert-xxlarge-v2& 0.18$\phantom{^{**}}$& 0.46$\phantom{^{**}}$& 0.95$^{*\phantom{*}}$\\\hline
roberta-base& -2.32$^{*\phantom{*}}$& 0.51$\phantom{^{**}}$& 0.36$\phantom{^{**}}$\\
roberta-large& -2.63$^{*\phantom{*}}$& 0.24$\phantom{^{**}}$& 0.82$^{*\phantom{*}}$\\\hline
google/electra-small-generator& -0.20$\phantom{^{**}}$& 0.71$^{*\phantom{*}}$& 0.85$^{*\phantom{*}}$\\
google/electra-large-generator& -2.64$^{*\phantom{*}}$& 0.73$^{*\phantom{*}}$& 0.63$^{*\phantom{*}}$\\ \hline
\end{tabular}
}
\caption{Three metrics of representational bias. Values are the average difference of associations between the target words ``he''/``she'', and a list of occupations as attributes. Larger values reflect a more severe bias. A positive value hints a skewed distribution towards males. A negative value hints a skewed distribution towards females. $*$: statistically significant at 0.01.}
\label{tab:bias-metrics}
\end{table}

\paragraph{Fairness} Our fairness evaluation is based on \citet{zhang-etal-2021-sociolectal}'s work, who study how the predictions of various PLMs align with the linguistic preferences of different social groups. They directly compare masked word predictions to human cloze tests, quantifying how often a language model agrees with the members of a particular social group on what is the most likely word in contexts such as: 

\begin{itemize}
    \item[] After waiting three hours, Cal whined and started to $[\mbox{MASK}]$.
\end{itemize}

\citet{zhang-etal-2021-sociolectal} use, as their fairness metric, the min-max difference in precision ($\mathrm{\Delta P@1}$) across groups defined by the cross-product of several protected attributes, including gender, age, race, and level of education. Since we are comparing with binary gender bias probes, we only consider fairness across (binary) gender here. 
We sample members of each group (female and male) in a balanced way across subgroups, as defined by the other variables. This is equivalent to reporting the macro-average across subgroups for each group. $\mathrm{\Delta P@1}$ is thus the difference in performance between male and female groups, macro-averaged across subgroups in the cloze test data. We follow \citet{zhang-etal-2021-sociolectal} in also reporting the difference in mean reciprocal rank as a second performance metric ($\mathrm{\Delta MRR}$). See the individual scores in Table~\ref{tab:sociolectal}.

\begin{table}[htbp!]
\centering
\resizebox{\columnwidth}{!}{
\begin{tabular}{m{0.50\linewidth}M{0.21\linewidth}M{0.21\linewidth}}
& $\Delta P@1$ & $\Delta MRR$\\ \hline
bert-base-uncased& 0.69& 1.57\\
bert-base-cased& 0.15& 0.74\\
bert-large-uncased& 0.91& 1.34\\
bert-large-cased& -0.07& 0.32\\
bert-base-multilingual-cased& 0.89& 0.54\\ \hline
distilbert-base-uncased& 1.63& 0.64\\
albert-base-v2& 0.74& 0.94\\
albert-large-v2& 1.45& 1.21\\
albert-xxlarge-v2& 0.48& 0.41\\ \hline
roberta-base& 0.14& 0.06\\
roberta-large& 0.68& 0.69\\ \hline
google/electra-small-generator& 0.97& 0.43\\
google/electra-large-generator& 1.22& 0.97\\
\end{tabular}
}
\caption{Macro-averaged precision and mean reciprocal rank differences between male and female subgroups following experiments in \cite{zhang-etal-2021-sociolectal}. Values close to zero are preferred for a more equitable model.}
\label{tab:sociolectal}
\end{table}

\begin{figure*}[htbp!]
    \centering 
\minipage{0.33\textwidth}
  \includegraphics[width=\linewidth]{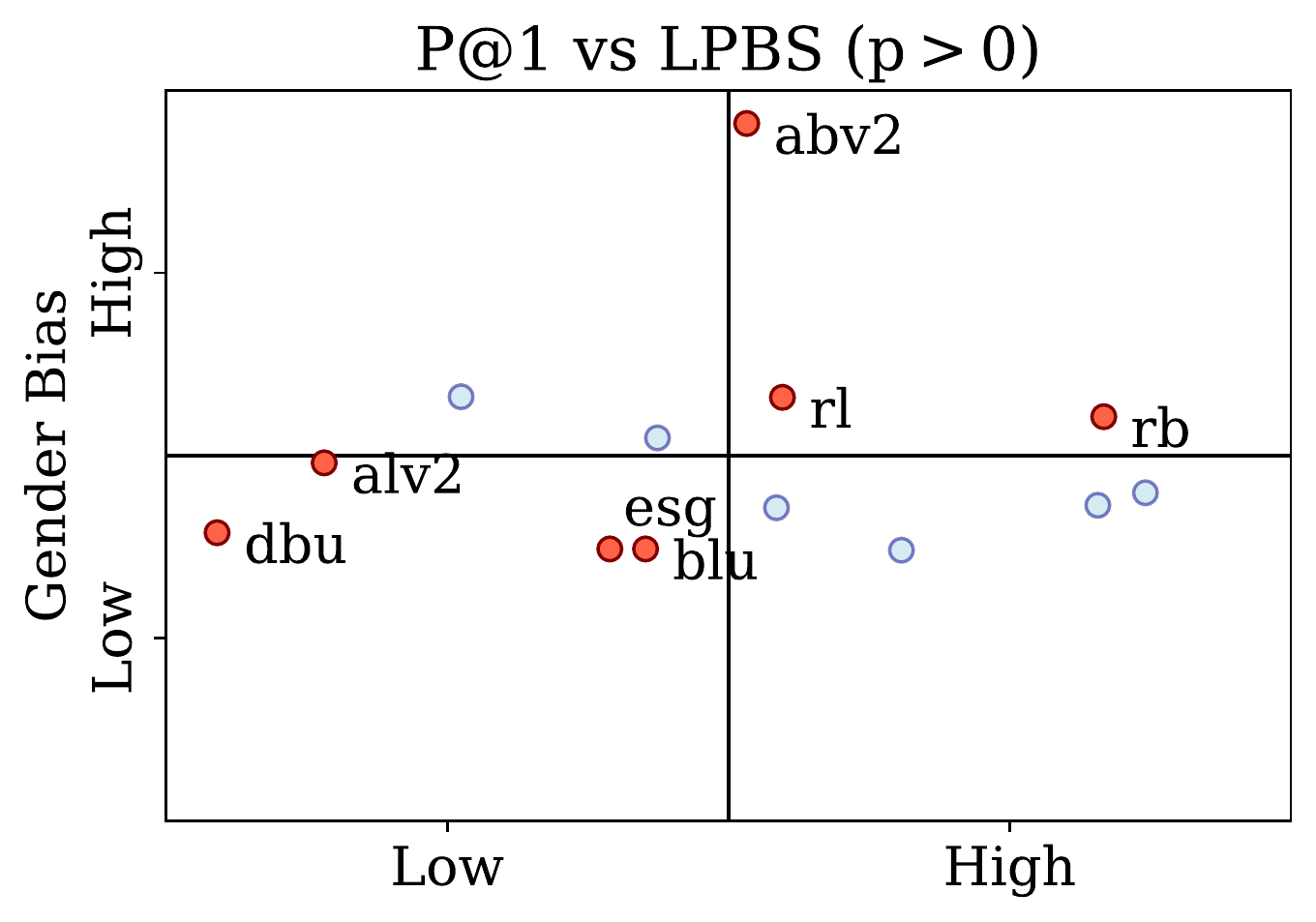}
\endminipage\hfill
\minipage{0.33\textwidth}
  \includegraphics[width=\linewidth]{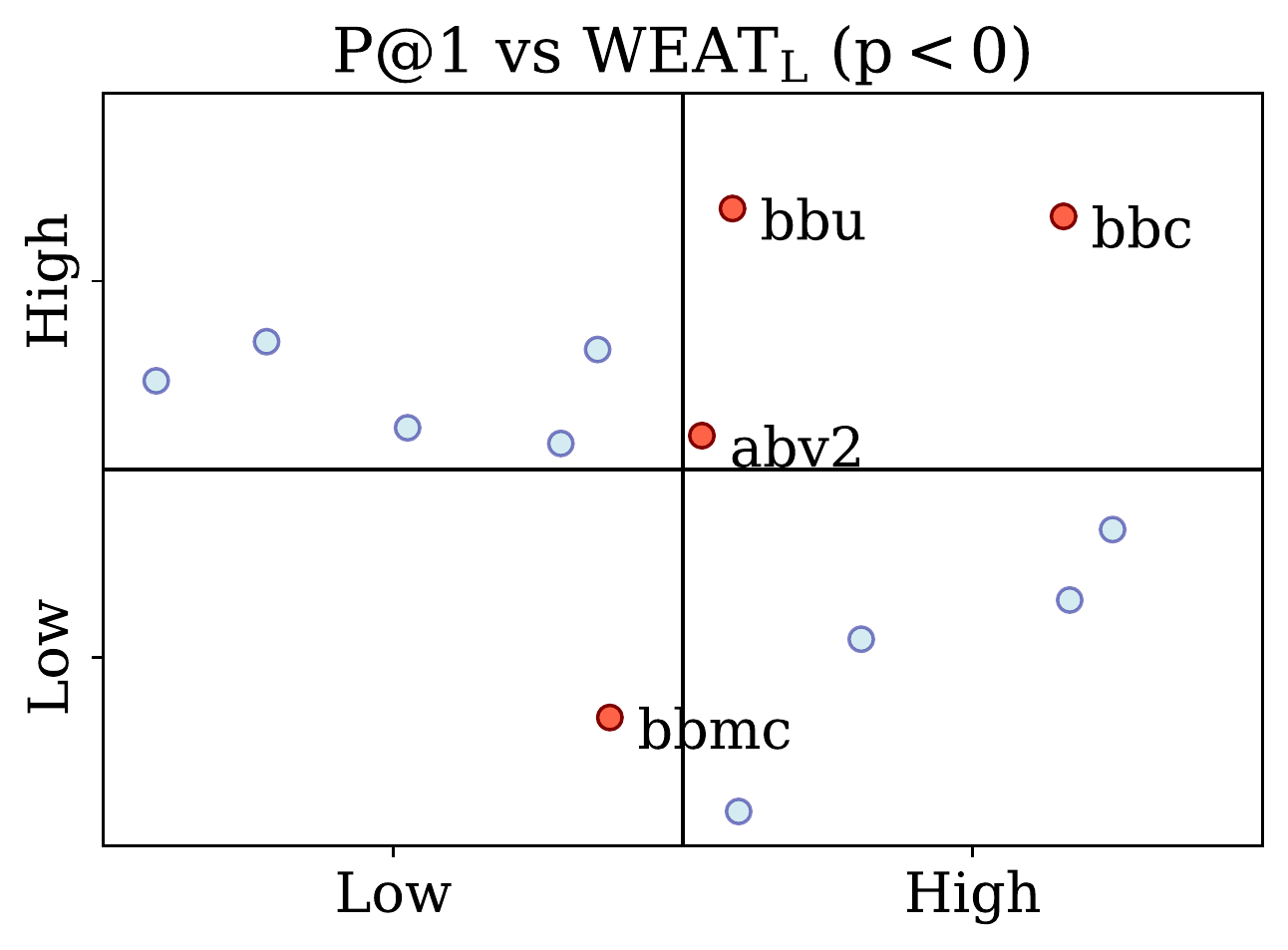}
\endminipage\hfill
\minipage{0.33\textwidth}
  \includegraphics[width=\linewidth]{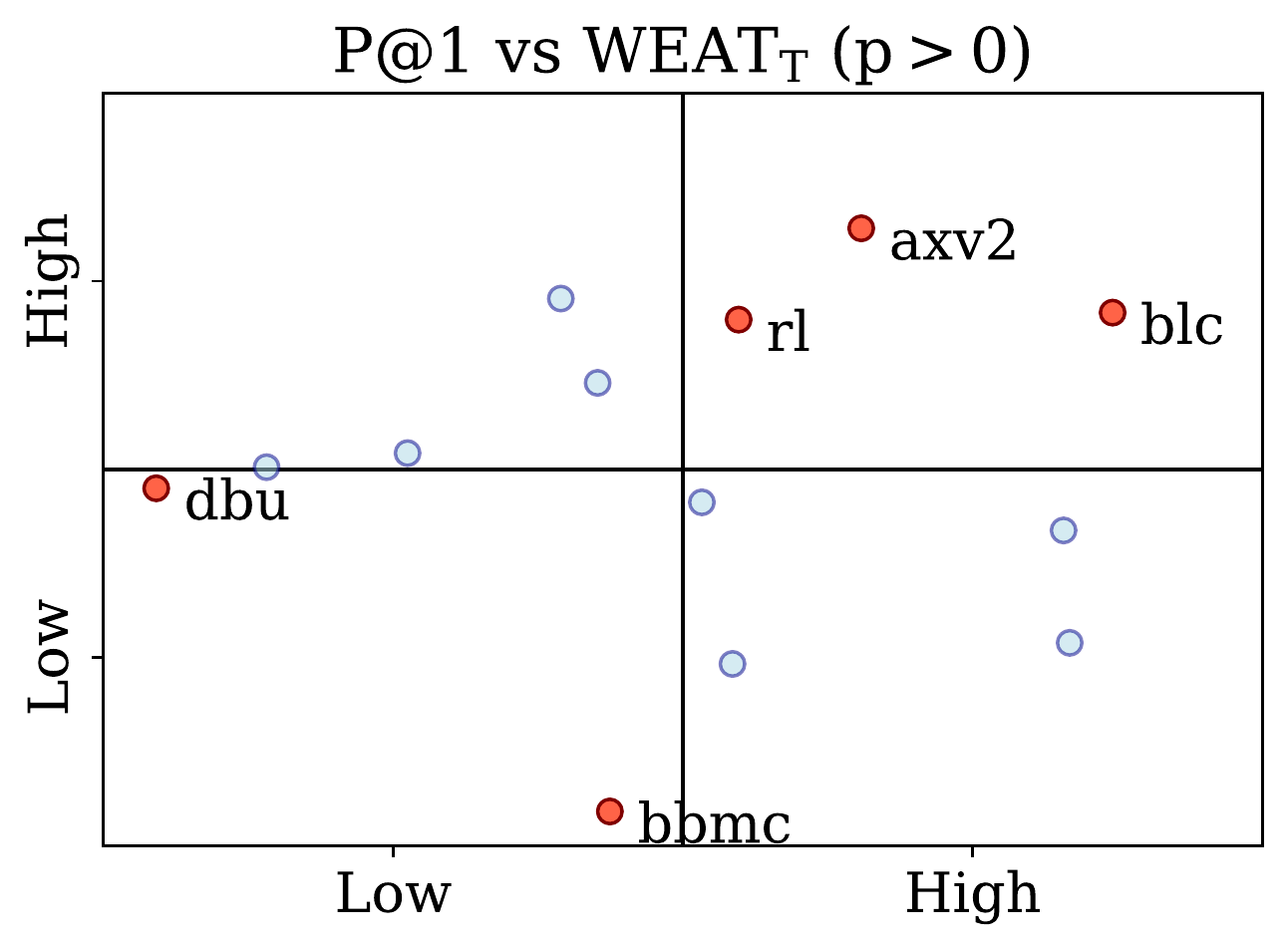}
\endminipage

\medskip
\minipage{0.33\textwidth}
  \includegraphics[width=\linewidth]{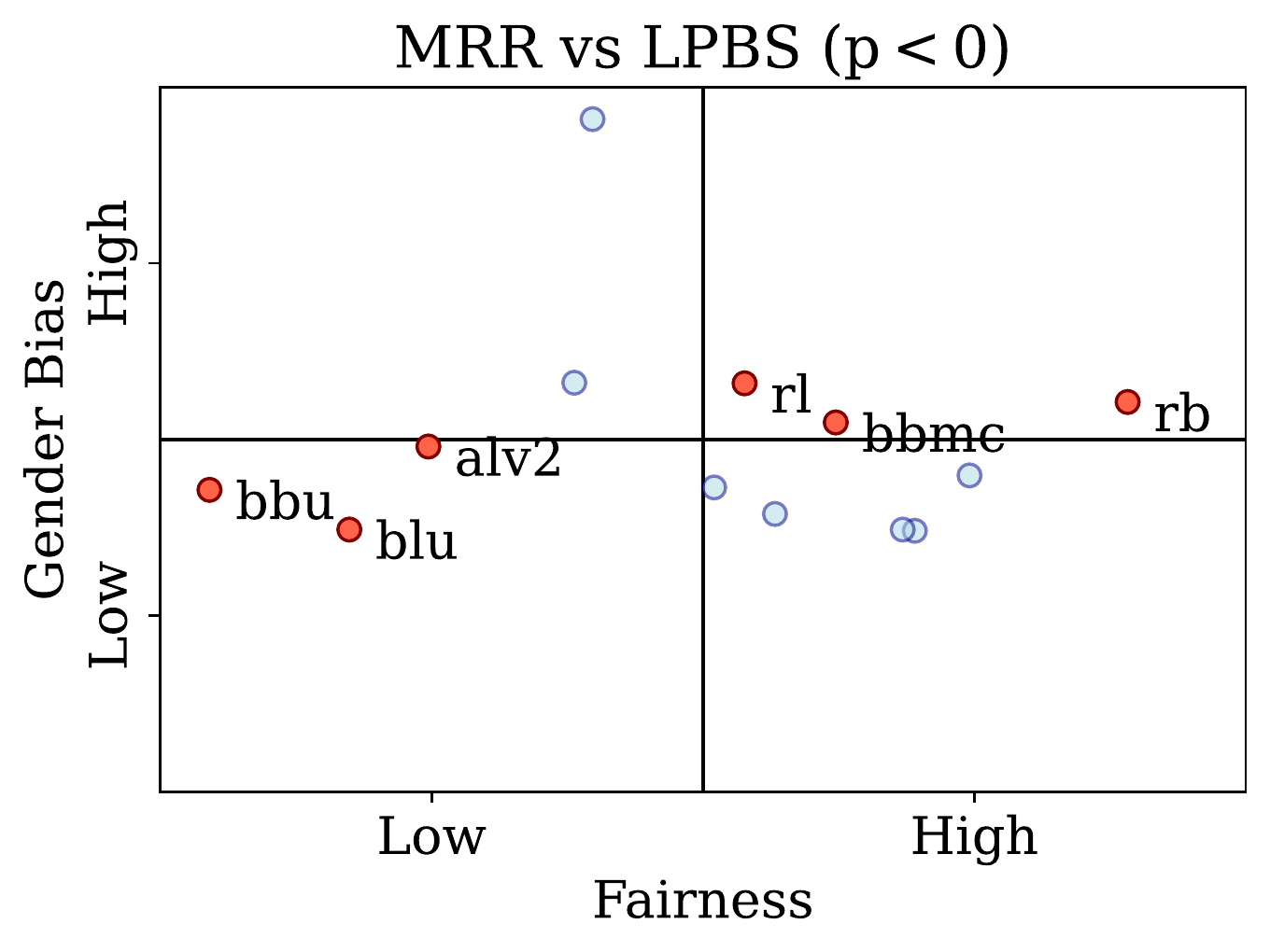}
\endminipage\hfill
\minipage{0.33\textwidth}
  \includegraphics[width=\linewidth]{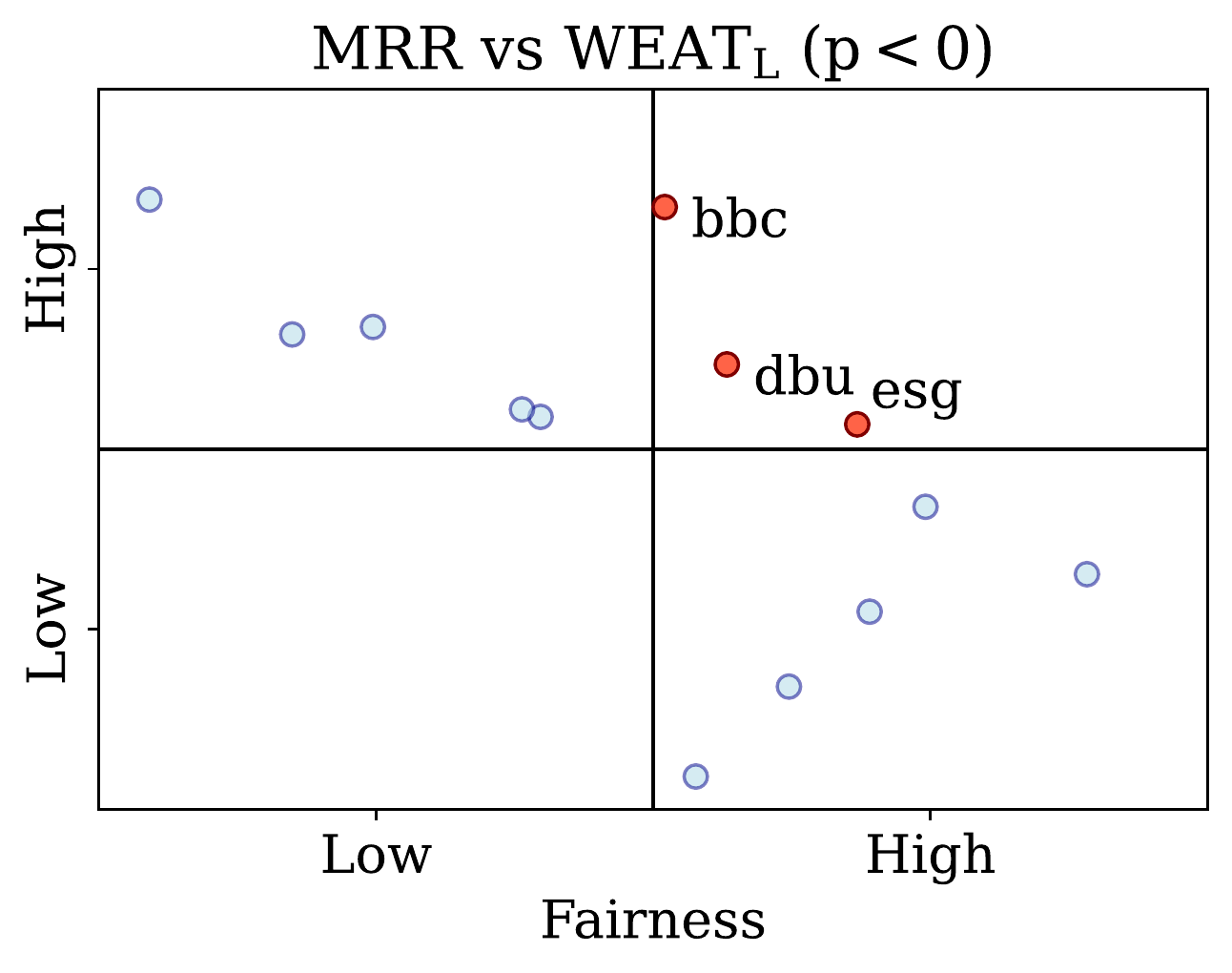}
\endminipage\hfill
\minipage{0.33\textwidth}
  \includegraphics[width=\linewidth]{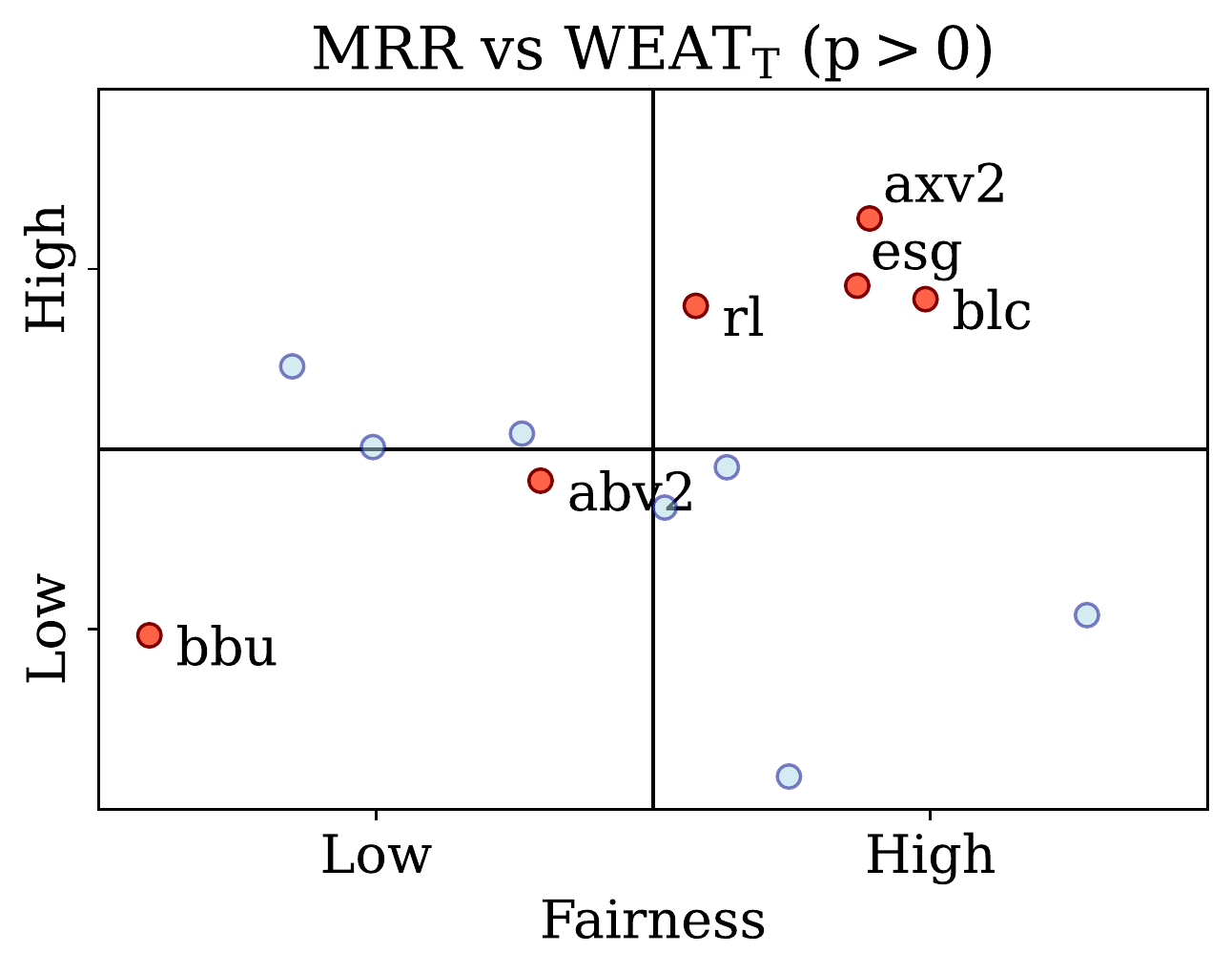}
\endminipage
\caption{Scatter plots show the relationship between different representational bias metrics and fairness evaluation. The upper row displays results when evaluating fairness through precision at top-1 ($\mathrm{P@1}$). The bottom row displays results when considering $\mathrm{MRR}$ to evaluate fairness. The division into quadrants is done according to average scores. Each point represents a language model, labelled with its initials. We see no support for a strong negative correlation between bias and fairness. Red points mark the clear counter-examples to such a negative correlation. Global trend for each plot is summarized with the sign of Pearson coefficient ($p$).}
\Description{Scatter plots show the relationship between different representational bias metrics and fairness evaluation. The upper row displays results when evaluating fairness through precision at top 1 (precision at 1). The bottom row displays results when considering MRR to evaluate fairness. The division into quadrants is done according to average scores. Each point represents a language model, labelled with its initials. We see no support for a strong negative correlation between bias and fairness. Red points mark the clear counter-examples to such a negative correlation. Global trend for each plot is summarized with the sign of Pearson coefficient p.}
\label{fig:axes}
\end{figure*}

Results show performance gaps between binary gender groups. Consequently, we would expect models exhibiting high degree of bias in Table~\ref{tab:bias-metrics} to be the least fair. However, this is not the case. Figure~\ref{fig:axes} displays the results for bias and fairness jointly, often highlighting the lack of correlation. Note that, ideally, all data-points should belong to the bottom-right quadrant.

\paragraph{Metrics are uncorrelated} Now that we have our evaluation framework defined, let us analyze whether representational bias correlates with outcome disparity. This amounts to studying the correlations between LPBS and WEAT metrics {and the min-max $\mathrm{P@1}$ difference across groups. We report the sign of the Pearson correlation coefficient to ease the interpretation of the (ideally) monotonic relationship}\footnote{We deliberately omit the magnitude of the Pearson coefficient to emphasize the sign of the correlation. Ideally, bias and fairness metrics should have a negative \emph{linear} dependence ($p<0$).} {between each set of metrics in Figure~\ref{fig:axes}}. 

Results are two-fold: 

\begin{itemize}
    \item[(i)] {The discrepancy across sub-graphs in Figure~\ref{fig:axes} aligns with results in \citet{may-etal-2019-measuring}, \citet{biased-rulers} and \citet{cao-etal-2022-intrinsic}, who all found different representational bias metrics to lead to mutually inconsistent results. WEAT$_L$ and WEAT$_T$ are related and show some agreement, but generally, results are wildly different across metrics.} 
    \item[(ii)] {More importantly, for our purposes, representational bias and fairness-as-equal-performance (quantified as min-max differences across performance scores for different groups) are, in fact, uncorrelated. }Models with high bias values are the most fair according to our fairness metric, and vice versa. These cases are highlighted in red in Figure~\ref{fig:axes}.
For example, {\tt roberta-base} ({\tt rb}) is among the most biased models according to LPBS, but it exhibits the highest degree of fairness wrt.~the $\mathrm{MRR}$ metric --and second highest wrt.~ $\mathrm{P@1}$. The bigger PLM, {\tt roberta-large} ({\tt rl}) is slightly less biased according to LPBS, but it is generally less fair. Values from the WEAT metrics are, in this case, somewhat mixed. 
\end{itemize}

\noindent Result (ii) is evidence {\em against}~the In-Group Affinity Assumption and {\em for}~the independence of bias and fairness.
Looking at each model family--separated by horizontal lines in Table~\ref{tab:bias-metrics} and \ref{tab:sociolectal}--model size does not systematically lead to larger or smaller bias scores, and it does not seem strongly correlated with any of the fairness metrics either.

In the following section, we survey research in the social sciences that also suggest the In-Group Affinity Assumption is mostly false, with one important caveat: Slur words have marked in-group usage. In most applications, this exception would be insufficient to drive a causal link between association bias and empirical fairness, because slur words are rare, and performance differences across social groups are pervasive.

\section{Association Bias and Empirical Fairness are Sometimes at Odds (in Humans)}
\label{sec:humans}

The thought experiment in \S\ref{sec:theory} shows that bias and fairness can in fact be completely independent or orthogonal.
The experiments in \S\ref{sec:practice} further showed that there is no direct correlation between the association bias in a model $\mathbf{M}$ toward social groups $g_1,\ldots,g_n$, and the performance disparity (fairness) of $\mathbf{M}$ across data from these groups $g_1,\ldots,g_n$.

In such cases, debiasing a model with respect to the representation of a certain group (e.g., $g_1$) has no impact on the performance of the model for users from the group. The beneficiaries of such a debiasing procedure are, in other words, not necessarily the group the debiasing was intended to increase fairness for. The idea that debiasing word representations that are related to a particular group increases the fairness of the model for that group, relies on the assumption that those words are also used by the in-group more frequently or in other ways than by other groups. This assumption--which we called the In-Group Affinity Assumption--seems problematic, since there are plenty of examples in the literature of the opposite. 
In the following, we briefly review some examples that originate from the NLP literature; others from the social sciences. 

We are often likely to talk more about members of other groups than our in-group peers. \citet{Li_Dickinson2017}, for example, find that some of the most indicative n-grams for detecting young female users on Chinese social media are the names of male pop stars. Correcting or debiasing the representations of these names would not improve model fairness on texts written by the male pop stars, but rather on texts written by young female users. 

\citet{MorganLopez2017PredictingAG} show that young (pre-college) children talk more about college on Twitter than adults in their college age. 

\citet{Wei_SantosJr2020} analyze data from Twitter and Reddit and find that the most predictive n-grams for Israeli users include ``Iraqis'' and ``Palestinians'', while for Palestinian users ``israeli military detention centres'' and ``Lieberman settler rabbis'' (referring to the Israeli Defence Minister, Avigdor Lieberman) are among the most predictive n-grams.

Generally, political debates are often experienced as negative in both tone and nature. According to a 2019 Pew Research Center study, 85\% of Americans say that the political debate has become ``more negative''.\footnote{\href{https://www.pewresearch.org/politics/2019/06/19/public-highly-critical-of-state-of-political-discourse-in-the-u-s/}{\nolinkurl{pewresearch.org/politics/2019/06/19/public-highly-critical-of-state-of-political-discourse-in-the-u-s/}}} One explanation for the increase in negative sentiment in political discourse is increased attention to what members of other (political) groups do wrong compared to what the in-group peers do right. 
Supporting this explanation, \citet{RePEc:bin:bpeajo:v:43:y:2012:i:2012-02:p:1-81} show, for example, that one of the most partisan phrases used by US Democrats in congressional texts was ``great Republican Party''. 

Similarly, \citet{Van_Duijnhoven2018} finds that Democrats on Twitter mention Trump and the Republican party more often than their Republican counterparts. In analyzing the language of German political parties, \citet{Biessmann2016} likewise finds that the left-wing party, Linke, has a high frequency of mentions of large corporations ({\em konzerne}) and policies that negatively impact the social welfare. 

\paragraph{On Slur}
Some slurring terms (e.g. ``dyke'', ``queer'' and ``bitch'') have been reclaimed or reappropriated by the target group resulting in a semantic discrepancy dependent on the speaker's group membership \cite{Ritchie2017-RITSII, HENRY2014185}. This results in what we term the {In-Group Affinity Assumption}, where the in-group's use of the term will differ significantly from that of the out-groups. Any debiasing of the term will have no significant impact on the performance for the in-groups, since the language model's representation of the term will reflect the majority use of the term, which will not be that of the in-group. However, since slurs are per definition defamatory terms, debiasing these terms will result in less insulting outputs in downstream tasks, and this may result in a higher perception of fairness for the target group.

\section{Discussion and Conclusion}

The independence of representational bias and fairness-as-equal-performance shown here, along with the falsification of the In-Group Affinity Assumption, runs counter to the NLP literature. Bias and fairness have been assumed to be intimately connected, and the In-Group Affinity Assumption has been implicit and unquestioned in much recent work. The results we present in this paper are, at the same time, in a sense not surprising. Or they {\em should} not be surprising. In many aspects of private and public life, we encounter decisions or patterns where bias and fairness exist or fluctuate independently of each other, or in which they are negatively correlated. In affirmative action, for example, we tolerate and encourage a (more) biased decision-making process to achieve (higher) fairness. While positive discrimination is heavily debated \cite{10.1257/jel.38.3.483, 10.2307/40388838, noon-positive-discrimination}, it is a good example of a biased process intended to increase the level of fairness. 



Methods for correctly assessing model biases remains an open research question. Current evaluation benchmarks give inconsistent results \cite{may-etal-2019-measuring,biased-rulers,cao-etal-2022-intrinsic}. Moreover, as discussed in \S\ref{sec2}, evaluating model biases with metrics that only consider local geometries, such as cosine-based metrics, can be inadequate. 
The fairness metric literature is also full of controversies \cite{Miconi2017TheIO,impossibility-friedler,kleinberg-trdeoffs,Hedden2021-HEDOSC}, but there is a broad consensus that performance disparity or outcome disparity is a real challenge for responsible NLP research and development. This consensus is not only limited to NLP research, but also found in legal studies, machine ethics, and the social sciences. Our results have shown that regardless of these open problems in bias and fairness research, the assumption that bias and fairness are always negatively correlated, and that one is a cause of the other, is not always true.  
Despite being closely related, it is important to understand that biases exist everywhere, but might not be unequivocally harmful. And similarly, fairness issues may arise in non-biased scenarios.

Finally, it is worth noting that we should not solely focus on the correlation between protected attributes such as race or gender and the model's output, but rather ask the question if \emph{they} are causing the outcome, and, whether the model is unfair to individuals in virtue of their membership in a certain group \cite{Hedden2021-HEDOSC}.

\paragraph{Conclusion} 
We reviewed part of the NLP literature showing how many researchers conflate bias and fairness, {i.e., representational bias and fairness-as-equal-performance,} or argue that fixing one will solve the other. 
In an attempt to explain why this does not hold always true, we devised a thought experiment in \S\ref{sec:theory}: a synthetic model that illustrates how bias and fairness can be completely independent of one another. We introduced the {In-Group Affinity Assumption} to highlight the assumption that a particular demographic groups use in-group terms more frequently--or in different ways--than other groups (non-standard). This, we argue, is a necessary assumption to drive a causal connection between bias and fairness, if it exists. 
In \S\ref{sec:humans}, we surveyed the social science literature and found evidence that often the opposite is the case, which substantiates our findings in \S\ref{sec:theory} and \S\ref{sec:practice}. Our survey includes examples from the social sciences, as well as from NLP research, where bias and fairness are (locally) {\em negatively} correlated. This provides strong reason to be skeptical of the In-Group Affinity Assumption and shows that bias and fairness are often independent or orthogonal to each other. 

In sum, we have shown the importance of studying bias and fairness independently of one another and cautioned against the In-Group Affinity Assumption. We think this, potentially, could lead to a valuable reorientation of the NLP literature, enabling researchers to study representational bias in more adequate ways, focusing on robustness and generation (to avoid bias reinforcement). This also highlights the different contributions of representational bias benchmarks and in-the-wild evaluation datasets with demographic information that can be used to evaluate performance disparities across groups. Bias and fairness seem to be separate issues, and we believe research should be done by disentangling the two.

\section{Limitations}
Our paper addresses the relationship between the two specific interpretations of bias and fairness, i.e., representational bias and fairness-as-equality. These are, in our view, the most common and most important definitions of bias and fairness in the NLP literature, but they are not the only ones. We hope others will follow up with studies of how other definitions relate. Our experiments in \S3~were limited to English benchmark datasets. We agree with \citet{ruder-etal-2022-square} that the prevalence of bias and fairness studies using English data, is most unfortunate, and we are, in parallel, working to create multilingual benchmarks for bias and fairness studies.

\bibliographystyle{ACM-Reference-Format}
\bibliography{sample-base}


\begin{thebibliography}{64}


\ifx \showCODEN    \undefined \def \showCODEN     #1{\unskip}     \fi
\ifx \showDOI      \undefined \def \showDOI       #1{#1}\fi
\ifx \showISBNx    \undefined \def \showISBNx     #1{\unskip}     \fi
\ifx \showISBNxiii \undefined \def \showISBNxiii  #1{\unskip}     \fi
\ifx \showISSN     \undefined \def \showISSN      #1{\unskip}     \fi
\ifx \showLCCN     \undefined \def \showLCCN      #1{\unskip}     \fi
\ifx \shownote     \undefined \def \shownote      #1{#1}          \fi
\ifx \showarticletitle \undefined \def \showarticletitle #1{#1}   \fi
\ifx \showURL      \undefined \def \showURL       {\relax}        \fi
\providecommand\bibfield[2]{#2}
\providecommand\bibinfo[2]{#2}
\providecommand\natexlab[1]{#1}
\providecommand\showeprint[2][]{arXiv:#2}

\bibitem[Ali et~al\mbox{.}(2022a)]%
        {DBLP:conf/icml/AliSEMMW22}
\bibfield{author}{\bibinfo{person}{Ameen Ali}, \bibinfo{person}{Thomas
  Schnake}, \bibinfo{person}{Oliver Eberle}, \bibinfo{person}{Gr{\'{e}}goire
  Montavon}, \bibinfo{person}{Klaus{-}Robert M{\"{u}}ller}, {and}
  \bibinfo{person}{Lior Wolf}.} \bibinfo{year}{2022}\natexlab{a}.
\newblock \showarticletitle{{XAI} for Transformers: Better Explanations through
  Conservative Propagation}. In \bibinfo{booktitle}{\emph{International
  Conference on Machine Learning, {ICML} 2022, 17-23 July 2022, Baltimore,
  Maryland, {USA}}} \emph{(\bibinfo{series}{Proceedings of Machine Learning
  Research}, Vol.~\bibinfo{volume}{162})},
  \bibfield{editor}{\bibinfo{person}{Kamalika Chaudhuri},
  \bibinfo{person}{Stefanie Jegelka}, \bibinfo{person}{Le~Song},
  \bibinfo{person}{Csaba Szepesv{\'{a}}ri}, \bibinfo{person}{Gang Niu}, {and}
  \bibinfo{person}{Sivan Sabato}} (Eds.). \bibinfo{publisher}{{PMLR}},
  \bibinfo{pages}{435--451}.
\newblock
\urldef\tempurl%
\url{https://proceedings.mlr.press/v162/ali22a.html}
\showURL{%
\tempurl}


\bibitem[Ali et~al\mbox{.}(2022b)]%
        {https://doi.org/10.48550/arxiv.2202.07304}
\bibfield{author}{\bibinfo{person}{Ameen Ali}, \bibinfo{person}{Thomas
  Schnake}, \bibinfo{person}{Oliver Eberle}, \bibinfo{person}{Grégoire
  Montavon}, \bibinfo{person}{Klaus-Robert Müller}, {and}
  \bibinfo{person}{Lior Wolf}.} \bibinfo{year}{2022}\natexlab{b}.
\newblock \bibinfo{title}{XAI for Transformers: Better Explanations through
  Conservative Propagation}.
\newblock
\newblock
\urldef\tempurl%
\url{https://doi.org/10.48550/ARXIV.2202.07304}
\showDOI{\tempurl}


\bibitem[Barmes(2009)]%
        {10.2307/40388838}
\bibfield{author}{\bibinfo{person}{Lizzie Barmes}.}
  \bibinfo{year}{2009}\natexlab{}.
\newblock \showarticletitle{Equality Law and Experimentation: The Positive
  Action Challenge}.
\newblock \bibinfo{journal}{\emph{The Cambridge Law Journal}}
  \bibinfo{volume}{68}, \bibinfo{number}{3} (\bibinfo{year}{2009}),
  \bibinfo{pages}{623--654}.
\newblock
\showISSN{00081973, 14692139}
\urldef\tempurl%
\url{http://www.jstor.org/stable/40388838}
\showURL{%
\tempurl}


\bibitem[Barocas et~al\mbox{.}(2019)]%
        {barocas-hardt-narayanan}
\bibfield{author}{\bibinfo{person}{Solon Barocas}, \bibinfo{person}{Moritz
  Hardt}, {and} \bibinfo{person}{Arvind Narayanan}.}
  \bibinfo{year}{2019}\natexlab{}.
\newblock \bibinfo{booktitle}{\emph{Fairness and Machine Learning}}.
\newblock \bibinfo{publisher}{fairmlbook.org}.
\newblock
\newblock
\shownote{\url{http://www.fairmlbook.org}}.


\bibitem[Bartl et~al\mbox{.}(2020)]%
        {bartl-etal-2020-unmasking}
\bibfield{author}{\bibinfo{person}{Marion Bartl}, \bibinfo{person}{Malvina
  Nissim}, {and} \bibinfo{person}{Albert Gatt}.}
  \bibinfo{year}{2020}\natexlab{}.
\newblock \showarticletitle{Unmasking Contextual Stereotypes: Measuring and
  Mitigating {BERT}{'}s Gender Bias}. In \bibinfo{booktitle}{\emph{Proceedings
  of the Second Workshop on Gender Bias in Natural Language Processing}}.
  \bibinfo{publisher}{Association for Computational Linguistics},
  \bibinfo{address}{Barcelona, Spain (Online)}, \bibinfo{pages}{1--16}.
\newblock
\urldef\tempurl%
\url{https://aclanthology.org/2020.gebnlp-1.1}
\showURL{%
\tempurl}


\bibitem[Bhatia(2017)]%
        {BHATIA201746}
\bibfield{author}{\bibinfo{person}{Sudeep Bhatia}.}
  \bibinfo{year}{2017}\natexlab{}.
\newblock \showarticletitle{The semantic representation of prejudice and
  stereotypes}.
\newblock \bibinfo{journal}{\emph{Cognition}}  \bibinfo{volume}{164}
  (\bibinfo{year}{2017}), \bibinfo{pages}{46--60}.
\newblock
\showISSN{0010-0277}
\urldef\tempurl%
\url{https://doi.org/10.1016/j.cognition.2017.03.016}
\showDOI{\tempurl}


\bibitem[Biessmann(2016)]%
        {Biessmann2016}
\bibfield{author}{\bibinfo{person}{Felix Biessmann}.}
  \bibinfo{year}{2016}\natexlab{}.
\newblock \bibinfo{title}{Automating political bias prediction}.
\newblock \bibinfo{howpublished}{arXiv preprint}.
\newblock
\newblock
\shownote{{a}rXiv:1608.02195}.


\bibitem[Blodgett et~al\mbox{.}(2020)]%
        {blodgett2020}
\bibfield{author}{\bibinfo{person}{Su~Lin Blodgett}, \bibinfo{person}{Solon
  Barocas}, \bibinfo{person}{Hal Daum{\'e}~III}, {and} \bibinfo{person}{Hanna
  Wallach}.} \bibinfo{year}{2020}\natexlab{}.
\newblock \showarticletitle{Language (Technology) is Power: A Critical Survey
  of {``}Bias{''} in {NLP}}. In \bibinfo{booktitle}{\emph{Proceedings of the
  58th Annual Meeting of the Association for Computational Linguistics}}.
  \bibinfo{publisher}{Association for Computational Linguistics},
  \bibinfo{address}{Online}, \bibinfo{pages}{5454--5476}.
\newblock
\urldef\tempurl%
\url{https://doi.org/10.18653/v1/2020.acl-main.485}
\showDOI{\tempurl}


\bibitem[Bolukbasi et~al\mbox{.}(2016)]%
        {bolukbasi2016man}
\bibfield{author}{\bibinfo{person}{Tolga Bolukbasi}, \bibinfo{person}{Kai-Wei
  Chang}, \bibinfo{person}{James~Y Zou}, \bibinfo{person}{Venkatesh Saligrama},
  {and} \bibinfo{person}{Adam~T Kalai}.} \bibinfo{year}{2016}\natexlab{}.
\newblock \showarticletitle{Man is to Computer Programmer as Woman is to
  Homemaker? Debiasing Word Embeddings}. In \bibinfo{booktitle}{\emph{Advances
  in Neural Information Processing Systems}},
  \bibfield{editor}{\bibinfo{person}{D.~Lee}, \bibinfo{person}{M.~Sugiyama},
  \bibinfo{person}{U.~Luxburg}, \bibinfo{person}{I.~Guyon}, {and}
  \bibinfo{person}{R.~Garnett}} (Eds.), Vol.~\bibinfo{volume}{29}.
  \bibinfo{publisher}{Curran Associates, Inc.}
\newblock
\urldef\tempurl%
\url{https://proceedings.neurips.cc/paper/2016/file/a486cd07e4ac3d270571622f4f316ec5-Paper.pdf}
\showURL{%
\tempurl}


\bibitem[Brunet et~al\mbox{.}(2019)]%
        {pmlr-v97-brunet19a}
\bibfield{author}{\bibinfo{person}{Marc-Etienne Brunet},
  \bibinfo{person}{Colleen Alkalay-Houlihan}, \bibinfo{person}{Ashton
  Anderson}, {and} \bibinfo{person}{Richard Zemel}.}
  \bibinfo{year}{2019}\natexlab{}.
\newblock \showarticletitle{Understanding the Origins of Bias in Word
  Embeddings}. In \bibinfo{booktitle}{\emph{Proceedings of the 36th
  International Conference on Machine Learning}}
  \emph{(\bibinfo{series}{Proceedings of Machine Learning Research},
  Vol.~\bibinfo{volume}{97})}, \bibfield{editor}{\bibinfo{person}{Kamalika
  Chaudhuri} {and} \bibinfo{person}{Ruslan Salakhutdinov}} (Eds.).
  \bibinfo{publisher}{PMLR}, \bibinfo{pages}{803--811}.
\newblock
\urldef\tempurl%
\url{https://proceedings.mlr.press/v97/brunet19a.html}
\showURL{%
\tempurl}


\bibitem[Caliskan et~al\mbox{.}(2022)]%
        {10.1145/3514094.3534162}
\bibfield{author}{\bibinfo{person}{Aylin Caliskan},
  \bibinfo{person}{Pimparkar~Parth Ajay}, \bibinfo{person}{Tessa Charlesworth},
  \bibinfo{person}{Robert Wolfe}, {and} \bibinfo{person}{Mahzarin~R. Banaji}.}
  \bibinfo{year}{2022}\natexlab{}.
\newblock \showarticletitle{Gender Bias in Word Embeddings: A Comprehensive
  Analysis of Frequency, Syntax, and Semantics}. In
  \bibinfo{booktitle}{\emph{Proceedings of the 2022 AAAI/ACM Conference on AI,
  Ethics, and Society}} (Oxford, United Kingdom) \emph{(\bibinfo{series}{AIES
  '22})}. \bibinfo{publisher}{Association for Computing Machinery},
  \bibinfo{address}{New York, NY, USA}, \bibinfo{pages}{156–170}.
\newblock
\showISBNx{9781450392471}
\urldef\tempurl%
\url{https://doi.org/10.1145/3514094.3534162}
\showDOI{\tempurl}


\bibitem[Caliskan et~al\mbox{.}(2017)]%
        {caliskan-weat}
\bibfield{author}{\bibinfo{person}{Aylin Caliskan}, \bibinfo{person}{Joanna~J.
  Bryson}, {and} \bibinfo{person}{Arvind Narayanan}.}
  \bibinfo{year}{2017}\natexlab{}.
\newblock \showarticletitle{Semantics derived automatically from language
  corpora contain human-like biases}.
\newblock \bibinfo{journal}{\emph{Science}} \bibinfo{volume}{356},
  \bibinfo{number}{6334} (\bibinfo{year}{2017}), \bibinfo{pages}{183--186}.
\newblock
\urldef\tempurl%
\url{https://doi.org/10.1126/science.aal4230}
\showDOI{\tempurl}
\showeprint{https://www.science.org/doi/pdf/10.1126/science.aal4230}


\bibitem[Cao et~al\mbox{.}(2022)]%
        {cao-etal-2022-intrinsic}
\bibfield{author}{\bibinfo{person}{Yang Cao}, \bibinfo{person}{Yada
  Pruksachatkun}, \bibinfo{person}{Kai-Wei Chang}, \bibinfo{person}{Rahul
  Gupta}, \bibinfo{person}{Varun Kumar}, \bibinfo{person}{Jwala Dhamala}, {and}
  \bibinfo{person}{Aram Galstyan}.} \bibinfo{year}{2022}\natexlab{}.
\newblock \showarticletitle{On the Intrinsic and Extrinsic Fairness Evaluation
  Metrics for Contextualized Language Representations}. In
  \bibinfo{booktitle}{\emph{Proceedings of the 60th Annual Meeting of the
  Association for Computational Linguistics (Volume 2: Short Papers)}}.
  \bibinfo{publisher}{Association for Computational Linguistics},
  \bibinfo{address}{Dublin, Ireland}, \bibinfo{pages}{561--570}.
\newblock
\urldef\tempurl%
\url{https://doi.org/10.18653/v1/2022.acl-short.62}
\showDOI{\tempurl}


\bibitem[Castelnovo et~al\mbox{.}(2022)]%
        {Castelnovo_2022}
\bibfield{author}{\bibinfo{person}{Alessandro Castelnovo},
  \bibinfo{person}{Riccardo Crupi}, \bibinfo{person}{Greta Greco},
  \bibinfo{person}{Daniele Regoli}, \bibinfo{person}{Ilaria~Giuseppina Penco},
  {and} \bibinfo{person}{Andrea~Claudio Cosentini}.}
  \bibinfo{year}{2022}\natexlab{}.
\newblock \showarticletitle{A clarification of the nuances in the fairness
  metrics landscape}.
\newblock \bibinfo{journal}{\emph{Scientific Reports}} \bibinfo{volume}{12},
  \bibinfo{number}{1} (\bibinfo{date}{March} \bibinfo{year}{2022}).
\newblock
\urldef\tempurl%
\url{https://doi.org/10.1038/s41598-022-07939-1}
\showDOI{\tempurl}


\bibitem[Chalkidis et~al\mbox{.}(2022)]%
        {chalkidis-etal-2022-fairlex}
\bibfield{author}{\bibinfo{person}{Ilias Chalkidis}, \bibinfo{person}{Tommaso
  Pasini}, \bibinfo{person}{Sheng Zhang}, \bibinfo{person}{Letizia Tomada},
  \bibinfo{person}{Sebastian Schwemer}, {and} \bibinfo{person}{Anders
  S{\o}gaard}.} \bibinfo{year}{2022}\natexlab{}.
\newblock \showarticletitle{{F}air{L}ex: A Multilingual Benchmark for
  Evaluating Fairness in Legal Text Processing}. In
  \bibinfo{booktitle}{\emph{Proceedings of the 60th Annual Meeting of the
  Association for Computational Linguistics (Volume 1: Long Papers)}}.
  \bibinfo{publisher}{Association for Computational Linguistics},
  \bibinfo{address}{Dublin, Ireland}, \bibinfo{pages}{4389--4406}.
\newblock
\urldef\tempurl%
\url{https://doi.org/10.18653/v1/2022.acl-long.301}
\showDOI{\tempurl}


\bibitem[Chaloner and Maldonado(2019)]%
        {chaloner-maldonado-2019-measuring}
\bibfield{author}{\bibinfo{person}{Kaytlin Chaloner} {and}
  \bibinfo{person}{Alfredo Maldonado}.} \bibinfo{year}{2019}\natexlab{}.
\newblock \showarticletitle{Measuring Gender Bias in Word Embeddings across
  Domains and Discovering New Gender Bias Word Categories}. In
  \bibinfo{booktitle}{\emph{Proceedings of the First Workshop on Gender Bias in
  Natural Language Processing}}. \bibinfo{publisher}{Association for
  Computational Linguistics}, \bibinfo{address}{Florence, Italy},
  \bibinfo{pages}{25--32}.
\newblock
\urldef\tempurl%
\url{https://doi.org/10.18653/v1/W19-3804}
\showDOI{\tempurl}


\bibitem[Chang et~al\mbox{.}(2019)]%
        {chang-etal-2019-bias}
\bibfield{author}{\bibinfo{person}{Kai-Wei Chang}, \bibinfo{person}{Vinodkumar
  Prabhakaran}, {and} \bibinfo{person}{Vicente Ordonez}.}
  \bibinfo{year}{2019}\natexlab{}.
\newblock \showarticletitle{Bias and Fairness in Natural Language Processing}.
  In \bibinfo{booktitle}{\emph{Proceedings of the 2019 Conference on Empirical
  Methods in Natural Language Processing and the 9th International Joint
  Conference on Natural Language Processing (EMNLP-IJCNLP): Tutorial
  Abstracts}}. \bibinfo{publisher}{Association for Computational Linguistics},
  \bibinfo{address}{Hong Kong, China}.
\newblock
\urldef\tempurl%
\url{https://aclanthology.org/D19-2004}
\showURL{%
\tempurl}


\bibitem[Chen et~al\mbox{.}(2020)]%
        {chen-etal-2020-analyzing}
\bibfield{author}{\bibinfo{person}{Wei-Fan Chen}, \bibinfo{person}{Khalid
  Al~Khatib}, \bibinfo{person}{Henning Wachsmuth}, {and} \bibinfo{person}{Benno
  Stein}.} \bibinfo{year}{2020}\natexlab{}.
\newblock \showarticletitle{Analyzing Political Bias and Unfairness in News
  Articles at Different Levels of Granularity}. In
  \bibinfo{booktitle}{\emph{Proceedings of the Fourth Workshop on Natural
  Language Processing and Computational Social Science}}.
  \bibinfo{publisher}{Association for Computational Linguistics},
  \bibinfo{address}{Online}, \bibinfo{pages}{149--154}.
\newblock
\urldef\tempurl%
\url{https://doi.org/10.18653/v1/2020.nlpcss-1.16}
\showDOI{\tempurl}


\bibitem[Crawford(2017)]%
        {crawford2017trouble}
\bibfield{author}{\bibinfo{person}{Kate Crawford}.}
  \bibinfo{year}{2017}\natexlab{}.
\newblock \showarticletitle{The trouble with bias}. In
  \bibinfo{booktitle}{\emph{Conference on {Neural} {Information} {Processing}
  {Systems}, invited speaker}}.
\newblock


\bibitem[Czarnowska et~al\mbox{.}(2021)]%
        {10.1162/tacl_a_00425}
\bibfield{author}{\bibinfo{person}{Paula Czarnowska}, \bibinfo{person}{Yogarshi
  Vyas}, {and} \bibinfo{person}{Kashif Shah}.} \bibinfo{year}{2021}\natexlab{}.
\newblock \showarticletitle{{Quantifying Social Biases in NLP: A Generalization
  and Empirical Comparison of Extrinsic Fairness Metrics}}.
\newblock \bibinfo{journal}{\emph{Transactions of the Association for
  Computational Linguistics}}  \bibinfo{volume}{9} (\bibinfo{date}{11}
  \bibinfo{year}{2021}), \bibinfo{pages}{1249--1267}.
\newblock
\showISSN{2307-387X}
\urldef\tempurl%
\url{https://doi.org/10.1162/tacl_a_00425}
\showDOI{\tempurl}
\showeprint{https://direct.mit.edu/tacl/article-pdf/doi/10.1162/tacl\_a\_00425/1972677/tacl\_a\_00425.pdf}


\bibitem[Dayanik and Pad{\'o}(2020)]%
        {dayanik-pado-2020-masking}
\bibfield{author}{\bibinfo{person}{Erenay Dayanik} {and}
  \bibinfo{person}{Sebastian Pad{\'o}}.} \bibinfo{year}{2020}\natexlab{}.
\newblock \showarticletitle{Masking Actor Information Leads to Fairer Political
  Claims Detection}. In \bibinfo{booktitle}{\emph{Proceedings of the 58th
  Annual Meeting of the Association for Computational Linguistics}}.
  \bibinfo{publisher}{Association for Computational Linguistics},
  \bibinfo{address}{Online}, \bibinfo{pages}{4385--4391}.
\newblock
\urldef\tempurl%
\url{https://doi.org/10.18653/v1/2020.acl-main.404}
\showDOI{\tempurl}


\bibitem[Delobelle et~al\mbox{.}(2021)]%
        {biased-rulers}
\bibfield{author}{\bibinfo{person}{Pieter Delobelle},
  \bibinfo{person}{Ewoenam~Kwaku Tokpo}, \bibinfo{person}{Toon Calders}, {and}
  \bibinfo{person}{Bettina Berendt}.} \bibinfo{year}{2021}\natexlab{}.
\newblock \showarticletitle{Measuring Fairness with Biased Rulers: A Survey on
  Quantifying Biases in Pretrained Language Models}.
\newblock \bibinfo{journal}{\emph{CoRR}}  \bibinfo{volume}{abs/2112.07447}
  (\bibinfo{year}{2021}).
\newblock
\urldef\tempurl%
\url{https://arxiv.org/abs/2112.07447}
\showURL{%
\tempurl}


\bibitem[Duijnhoven(2018)]%
        {Van_Duijnhoven2018}
\bibfield{author}{\bibinfo{person}{Coen~Van Duijnhoven}.}
  \bibinfo{year}{2018}\natexlab{}.
\newblock \emph{\bibinfo{title}{Predicting political preference through
  content- and stylistic text features and distant labeling}}.
\newblock \bibinfo{thesistype}{Master's\ thesis}. \bibinfo{school}{Tilburg
  University}.
\newblock


\bibitem[Friedler et~al\mbox{.}(2016)]%
        {impossibility-friedler}
\bibfield{author}{\bibinfo{person}{Sorelle~A. Friedler},
  \bibinfo{person}{Carlos Scheidegger}, {and} \bibinfo{person}{Suresh
  Venkatasubramanian}.} \bibinfo{year}{2016}\natexlab{}.
\newblock \bibinfo{title}{On the (im)possibility of fairness}.
\newblock
\newblock
\urldef\tempurl%
\url{https://doi.org/10.48550/ARXIV.1609.07236}
\showDOI{\tempurl}


\bibitem[Friedrich et~al\mbox{.}(2021)]%
        {friedrich-etal-2021-debie}
\bibfield{author}{\bibinfo{person}{Niklas Friedrich}, \bibinfo{person}{Anne
  Lauscher}, \bibinfo{person}{Simone~Paolo Ponzetto}, {and}
  \bibinfo{person}{Goran Glava{\v{s}}}.} \bibinfo{year}{2021}\natexlab{}.
\newblock \showarticletitle{{D}eb{IE}: A Platform for Implicit and Explicit
  Debiasing of Word Embedding Spaces}. In \bibinfo{booktitle}{\emph{Proceedings
  of the 16th Conference of the European Chapter of the Association for
  Computational Linguistics: System Demonstrations}}.
  \bibinfo{publisher}{Association for Computational Linguistics},
  \bibinfo{address}{Online}, \bibinfo{pages}{91--98}.
\newblock
\urldef\tempurl%
\url{https://doi.org/10.18653/v1/2021.eacl-demos.11}
\showDOI{\tempurl}


\bibitem[Goldfarb-Tarrant et~al\mbox{.}(2021)]%
        {goldfarb-tarrant-etal-2021-intrinsic}
\bibfield{author}{\bibinfo{person}{Seraphina Goldfarb-Tarrant},
  \bibinfo{person}{Rebecca Marchant}, \bibinfo{person}{Ricardo
  Mu{\~n}oz~S{\'a}nchez}, \bibinfo{person}{Mugdha Pandya}, {and}
  \bibinfo{person}{Adam Lopez}.} \bibinfo{year}{2021}\natexlab{}.
\newblock \showarticletitle{Intrinsic Bias Metrics Do Not Correlate with
  Application Bias}. In \bibinfo{booktitle}{\emph{Proceedings of the 59th
  Annual Meeting of the Association for Computational Linguistics and the 11th
  International Joint Conference on Natural Language Processing (Volume 1: Long
  Papers)}}. \bibinfo{publisher}{Association for Computational Linguistics},
  \bibinfo{address}{Online}, \bibinfo{pages}{1926--1940}.
\newblock
\urldef\tempurl%
\url{https://doi.org/10.18653/v1/2021.acl-long.150}
\showDOI{\tempurl}


\bibitem[Gonen and Goldberg(2019)]%
        {gonen-goldberg-2019-lipstick}
\bibfield{author}{\bibinfo{person}{Hila Gonen} {and} \bibinfo{person}{Yoav
  Goldberg}.} \bibinfo{year}{2019}\natexlab{}.
\newblock \showarticletitle{Lipstick on a Pig: {D}ebiasing Methods Cover up
  Systematic Gender Biases in Word Embeddings But do not Remove Them}. In
  \bibinfo{booktitle}{\emph{Proceedings of the 2019 Conference of the North
  {A}merican Chapter of the Association for Computational Linguistics: Human
  Language Technologies, Volume 1 (Long and Short Papers)}}.
  \bibinfo{publisher}{Association for Computational Linguistics},
  \bibinfo{address}{Minneapolis, Minnesota}, \bibinfo{pages}{609--614}.
\newblock
\urldef\tempurl%
\url{https://doi.org/10.18653/v1/N19-1061}
\showDOI{\tempurl}


\bibitem[Hansen and S{\o}gaard(2021)]%
        {hansen-sogaard-2021-lottery}
\bibfield{author}{\bibinfo{person}{Victor Petr{\'e}n~Bach Hansen} {and}
  \bibinfo{person}{Anders S{\o}gaard}.} \bibinfo{year}{2021}\natexlab{}.
\newblock \showarticletitle{Is the Lottery Fair? Evaluating Winning Tickets
  Across Demographics}. In \bibinfo{booktitle}{\emph{Findings of the
  Association for Computational Linguistics: ACL-IJCNLP 2021}}.
  \bibinfo{publisher}{Association for Computational Linguistics},
  \bibinfo{address}{Online}, \bibinfo{pages}{3214--3224}.
\newblock
\urldef\tempurl%
\url{https://doi.org/10.18653/v1/2021.findings-acl.284}
\showDOI{\tempurl}


\bibitem[Hashimoto et~al\mbox{.}(2018)]%
        {DBLP:conf/icml/HashimotoSNL18}
\bibfield{author}{\bibinfo{person}{Tatsunori~B. Hashimoto},
  \bibinfo{person}{Megha Srivastava}, \bibinfo{person}{Hongseok Namkoong},
  {and} \bibinfo{person}{Percy Liang}.} \bibinfo{year}{2018}\natexlab{}.
\newblock \showarticletitle{Fairness Without Demographics in Repeated Loss
  Minimization}. In \bibinfo{booktitle}{\emph{Proceedings of the 35th
  International Conference on Machine Learning, {ICML} 2018,
  Stockholmsm{\"{a}}ssan, Stockholm, Sweden, July 10-15, 2018}}
  \emph{(\bibinfo{series}{Proceedings of Machine Learning Research},
  Vol.~\bibinfo{volume}{80})}, \bibfield{editor}{\bibinfo{person}{Jennifer~G.
  Dy} {and} \bibinfo{person}{Andreas Krause}} (Eds.).
  \bibinfo{publisher}{{PMLR}}, \bibinfo{pages}{1934--1943}.
\newblock
\urldef\tempurl%
\url{http://proceedings.mlr.press/v80/hashimoto18a.html}
\showURL{%
\tempurl}


\bibitem[Hedden(2021)]%
        {Hedden2021-HEDOSC}
\bibfield{author}{\bibinfo{person}{Brian Hedden}.}
  \bibinfo{year}{2021}\natexlab{}.
\newblock \showarticletitle{On Statistical Criteria of Algorithmic Fairness}.
\newblock \bibinfo{journal}{\emph{Philosophy and Public Affairs}}
  \bibinfo{volume}{49}, \bibinfo{number}{2} (\bibinfo{year}{2021}),
  \bibinfo{pages}{209--231}.
\newblock
\urldef\tempurl%
\url{https://doi.org/10.1111/papa.12189}
\showDOI{\tempurl}


\bibitem[Henry et~al\mbox{.}(2014)]%
        {HENRY2014185}
\bibfield{author}{\bibinfo{person}{P.J. Henry}, \bibinfo{person}{Sarah~E.
  Butler}, {and} \bibinfo{person}{Mark~J. Brandt}.}
  \bibinfo{year}{2014}\natexlab{}.
\newblock \showarticletitle{The influence of target group status on the
  perception of the offensiveness of group-based slurs}.
\newblock \bibinfo{journal}{\emph{Journal of Experimental Social Psychology}}
  \bibinfo{volume}{53} (\bibinfo{year}{2014}), \bibinfo{pages}{185--192}.
\newblock
\showISSN{0022-1031}
\urldef\tempurl%
\url{https://doi.org/10.1016/j.jesp.2014.03.012}
\showDOI{\tempurl}


\bibitem[Holzer and Neumark(2000)]%
        {10.1257/jel.38.3.483}
\bibfield{author}{\bibinfo{person}{Harry Holzer} {and} \bibinfo{person}{David
  Neumark}.} \bibinfo{year}{2000}\natexlab{}.
\newblock \showarticletitle{Assessing Affirmative Action}.
\newblock \bibinfo{journal}{\emph{Journal of Economic Literature}}
  \bibinfo{volume}{38}, \bibinfo{number}{3} (\bibinfo{date}{September}
  \bibinfo{year}{2000}), \bibinfo{pages}{483--568}.
\newblock
\urldef\tempurl%
\url{https://doi.org/10.1257/jel.38.3.483}
\showDOI{\tempurl}


\bibitem[Hutchinson et~al\mbox{.}(2020)]%
        {hutchinson-etal-2020-social}
\bibfield{author}{\bibinfo{person}{Ben Hutchinson}, \bibinfo{person}{Vinodkumar
  Prabhakaran}, \bibinfo{person}{Emily Denton}, \bibinfo{person}{Kellie
  Webster}, \bibinfo{person}{Yu Zhong}, {and} \bibinfo{person}{Stephen
  Denuyl}.} \bibinfo{year}{2020}\natexlab{}.
\newblock \showarticletitle{Social Biases in {NLP} Models as Barriers for
  Persons with Disabilities}. In \bibinfo{booktitle}{\emph{Proceedings of the
  58th Annual Meeting of the Association for Computational Linguistics}}.
  \bibinfo{publisher}{Association for Computational Linguistics},
  \bibinfo{address}{Online}, \bibinfo{pages}{5491--5501}.
\newblock
\urldef\tempurl%
\url{https://doi.org/10.18653/v1/2020.acl-main.487}
\showDOI{\tempurl}


\bibitem[Jensen et~al\mbox{.}(2012)]%
        {RePEc:bin:bpeajo:v:43:y:2012:i:2012-02:p:1-81}
\bibfield{author}{\bibinfo{person}{Jacob Jensen}, \bibinfo{person}{Ethan
  Kaplan}, \bibinfo{person}{Suresh Naidu}, {and} \bibinfo{person}{Laurence
  Wilse-Samson}.} \bibinfo{year}{2012}\natexlab{}.
\newblock \showarticletitle{{Political Polarization and the Dynamics of
  Political Language: Evidence from 130 Years of Partisan Speech}}.
\newblock \bibinfo{journal}{\emph{Brookings Papers on Economic Activity}}
  \bibinfo{volume}{43}, \bibinfo{number}{2 (Fall)} (\bibinfo{year}{2012}),
  \bibinfo{pages}{1--81}.
\newblock
\urldef\tempurl%
\url{https://ideas.repec.org/a/bin/bpeajo/v43y2012i2012-02p1-81.html}
\showURL{%
\tempurl}


\bibitem[Kaneko et~al\mbox{.}(2022)]%
        {kaneko-etal-2022-debiasing}
\bibfield{author}{\bibinfo{person}{Masahiro Kaneko}, \bibinfo{person}{Danushka
  Bollegala}, {and} \bibinfo{person}{Naoaki Okazaki}.}
  \bibinfo{year}{2022}\natexlab{}.
\newblock \showarticletitle{Debiasing Isn{'}t Enough! {--} on the Effectiveness
  of Debiasing {MLM}s and Their Social Biases in Downstream Tasks}. In
  \bibinfo{booktitle}{\emph{Proceedings of the 29th International Conference on
  Computational Linguistics}}. \bibinfo{publisher}{International Committee on
  Computational Linguistics}, \bibinfo{address}{Gyeongju, Republic of Korea},
  \bibinfo{pages}{1299--1310}.
\newblock
\urldef\tempurl%
\url{https://aclanthology.org/2022.coling-1.111}
\showURL{%
\tempurl}


\bibitem[Kleinberg et~al\mbox{.}(2016)]%
        {kleinberg-trdeoffs}
\bibfield{author}{\bibinfo{person}{Jon Kleinberg}, \bibinfo{person}{Sendhil
  Mullainathan}, {and} \bibinfo{person}{Manish Raghavan}.}
  \bibinfo{year}{2016}\natexlab{}.
\newblock \bibinfo{title}{Inherent Trade-Offs in the Fair Determination of Risk
  Scores}.
\newblock
\newblock
\urldef\tempurl%
\url{https://doi.org/10.48550/ARXIV.1609.05807}
\showDOI{\tempurl}


\bibitem[Kurita et~al\mbox{.}(2019)]%
        {kurita-etal-2019-measuring}
\bibfield{author}{\bibinfo{person}{Keita Kurita}, \bibinfo{person}{Nidhi Vyas},
  \bibinfo{person}{Ayush Pareek}, \bibinfo{person}{Alan~W Black}, {and}
  \bibinfo{person}{Yulia Tsvetkov}.} \bibinfo{year}{2019}\natexlab{}.
\newblock \showarticletitle{Measuring Bias in Contextualized Word
  Representations}. In \bibinfo{booktitle}{\emph{Proceedings of the First
  Workshop on Gender Bias in Natural Language Processing}}.
  \bibinfo{publisher}{Association for Computational Linguistics},
  \bibinfo{address}{Florence, Italy}, \bibinfo{pages}{166--172}.
\newblock
\urldef\tempurl%
\url{https://doi.org/10.18653/v1/W19-3823}
\showDOI{\tempurl}


\bibitem[Lauscher et~al\mbox{.}(2021)]%
        {lauscher-etal-2021-sustainable-modular}
\bibfield{author}{\bibinfo{person}{Anne Lauscher}, \bibinfo{person}{Tobias
  Lueken}, {and} \bibinfo{person}{Goran Glava{\v{s}}}.}
  \bibinfo{year}{2021}\natexlab{}.
\newblock \showarticletitle{Sustainable Modular Debiasing of Language Models}.
  In \bibinfo{booktitle}{\emph{Findings of the Association for Computational
  Linguistics: EMNLP 2021}}. \bibinfo{publisher}{Association for Computational
  Linguistics}, \bibinfo{address}{Punta Cana, Dominican Republic},
  \bibinfo{pages}{4782--4797}.
\newblock
\urldef\tempurl%
\url{https://doi.org/10.18653/v1/2021.findings-emnlp.411}
\showDOI{\tempurl}


\bibitem[Li and Dickinson(2017)]%
        {Li_Dickinson2017}
\bibfield{author}{\bibinfo{person}{Wen Li} {and} \bibinfo{person}{Markus
  Dickinson}.} \bibinfo{year}{2017}\natexlab{}.
\newblock \showarticletitle{Gender Prediction for Chinese Social Media Data}.
  In \bibinfo{booktitle}{\emph{Proceedings of Recent Advances in Natural
  Language Processing}}.
\newblock


\bibitem[Liu et~al\mbox{.}(2020)]%
        {liu-etal-2020-gender}
\bibfield{author}{\bibinfo{person}{Haochen Liu}, \bibinfo{person}{Jamell
  Dacon}, \bibinfo{person}{Wenqi Fan}, \bibinfo{person}{Hui Liu},
  \bibinfo{person}{Zitao Liu}, {and} \bibinfo{person}{Jiliang Tang}.}
  \bibinfo{year}{2020}\natexlab{}.
\newblock \showarticletitle{Does Gender Matter? Towards Fairness in Dialogue
  Systems}. In \bibinfo{booktitle}{\emph{Proceedings of the 28th International
  Conference on Computational Linguistics}}. \bibinfo{publisher}{International
  Committee on Computational Linguistics}, \bibinfo{address}{Barcelona, Spain
  (Online)}, \bibinfo{pages}{4403--4416}.
\newblock
\urldef\tempurl%
\url{https://doi.org/10.18653/v1/2020.coling-main.390}
\showDOI{\tempurl}


\bibitem[Maity et~al\mbox{.}(2020)]%
        {subpopulation-shift-fairness}
\bibfield{author}{\bibinfo{person}{Subha Maity}, \bibinfo{person}{Debarghya
  Mukherjee}, \bibinfo{person}{Mikhail Yurochkin}, {and}
  \bibinfo{person}{Yuekai Sun}.} \bibinfo{year}{2020}\natexlab{}.
\newblock \bibinfo{title}{Does enforcing fairness mitigate biases caused by
  subpopulation shift?}
\newblock
\newblock
\urldef\tempurl%
\url{https://doi.org/10.48550/ARXIV.2011.03173}
\showDOI{\tempurl}


\bibitem[May et~al\mbox{.}(2019)]%
        {may-etal-2019-measuring}
\bibfield{author}{\bibinfo{person}{Chandler May}, \bibinfo{person}{Alex Wang},
  \bibinfo{person}{Shikha Bordia}, \bibinfo{person}{Samuel~R. Bowman}, {and}
  \bibinfo{person}{Rachel Rudinger}.} \bibinfo{year}{2019}\natexlab{}.
\newblock \showarticletitle{On Measuring Social Biases in Sentence Encoders}.
  In \bibinfo{booktitle}{\emph{Proceedings of the 2019 Conference of the North
  {A}merican Chapter of the Association for Computational Linguistics: Human
  Language Technologies, Volume 1 (Long and Short Papers)}}.
  \bibinfo{publisher}{Association for Computational Linguistics},
  \bibinfo{address}{Minneapolis, Minnesota}, \bibinfo{pages}{622--628}.
\newblock
\urldef\tempurl%
\url{https://doi.org/10.18653/v1/N19-1063}
\showDOI{\tempurl}


\bibitem[Mehrabi et~al\mbox{.}(2021)]%
        {mehrabi-survey}
\bibfield{author}{\bibinfo{person}{Ninareh Mehrabi}, \bibinfo{person}{Fred
  Morstatter}, \bibinfo{person}{Nripsuta Saxena}, \bibinfo{person}{Kristina
  Lerman}, {and} \bibinfo{person}{Aram Galstyan}.}
  \bibinfo{year}{2021}\natexlab{}.
\newblock \showarticletitle{A Survey on Bias and Fairness in Machine Learning}.
\newblock \bibinfo{journal}{\emph{ACM Comput. Surv.}} \bibinfo{volume}{54},
  \bibinfo{number}{6}, Article \bibinfo{articleno}{115} (\bibinfo{date}{jul}
  \bibinfo{year}{2021}), \bibinfo{numpages}{35}~pages.
\newblock
\showISSN{0360-0300}
\urldef\tempurl%
\url{https://doi.org/10.1145/3457607}
\showDOI{\tempurl}


\bibitem[Miconi(2017)]%
        {Miconi2017TheIO}
\bibfield{author}{\bibinfo{person}{Thomas Miconi}.}
  \bibinfo{year}{2017}\natexlab{}.
\newblock \showarticletitle{The impossibility of "fairness": a generalized
  impossibility result for decisions}.
\newblock \bibinfo{journal}{\emph{arXiv: Applications}} (\bibinfo{year}{2017}).
\newblock
\urldef\tempurl%
\url{https://doi.org/10.48550/ARXIV.1707.01195}
\showDOI{\tempurl}


\bibitem[Morgan-Lopez et~al\mbox{.}(2017)]%
        {MorganLopez2017PredictingAG}
\bibfield{author}{\bibinfo{person}{Antonio~Alexander Morgan-Lopez},
  \bibinfo{person}{Annice~E Kim}, \bibinfo{person}{Robert~F. Chew}, {and}
  \bibinfo{person}{Paul Ruddle}.} \bibinfo{year}{2017}\natexlab{}.
\newblock \showarticletitle{Predicting age groups of Twitter users based on
  language and metadata features}.
\newblock \bibinfo{journal}{\emph{PLoS ONE}}  \bibinfo{volume}{12}
  (\bibinfo{year}{2017}).
\newblock


\bibitem[Noon(2010)]%
        {noon-positive-discrimination}
\bibfield{author}{\bibinfo{person}{Mike Noon}.}
  \bibinfo{year}{2010}\natexlab{}.
\newblock \showarticletitle{The shackled runner: time to rethink positive
  discrimination?}
\newblock \bibinfo{journal}{\emph{Work, Employment and Society}}
  \bibinfo{volume}{24}, \bibinfo{number}{4} (\bibinfo{year}{2010}),
  \bibinfo{pages}{728--739}.
\newblock
\urldef\tempurl%
\url{https://doi.org/10.1177/0950017010380648}
\showDOI{\tempurl}
\showeprint{https://doi.org/10.1177/0950017010380648}


\bibitem[Qian et~al\mbox{.}(2022)]%
        {qian2022perturbation}
\bibfield{author}{\bibinfo{person}{Rebecca Qian}, \bibinfo{person}{Candace
  Ross}, \bibinfo{person}{Jude Fernandes}, \bibinfo{person}{Eric Smith},
  \bibinfo{person}{Douwe Kiela}, {and} \bibinfo{person}{Adina Williams}.}
  \bibinfo{year}{2022}\natexlab{}.
\newblock \bibinfo{title}{Perturbation Augmentation for Fairer NLP}.
\newblock
\newblock
\urldef\tempurl%
\url{https://doi.org/10.48550/ARXIV.2205.12586}
\showDOI{\tempurl}


\bibitem[Rawls(1971)]%
        {rawls_theory_1971}
\bibfield{author}{\bibinfo{person}{John Rawls}.}
  \bibinfo{year}{1971}\natexlab{}.
\newblock \bibinfo{booktitle}{\emph{A Theory of Justice} (\bibinfo{edition}{1}
  ed.)}.
\newblock \bibinfo{publisher}{Belknap Press of Harvard University Press},
  \bibinfo{address}{Cambridge, Massachussets}.
\newblock
\showISBNx{0-674-88014-5}


\bibitem[Reddy et~al\mbox{.}(2021)]%
        {Benchmarking2021_2723d092}
\bibfield{author}{\bibinfo{person}{Charan Reddy}, \bibinfo{person}{Deepak
  Sharma}, \bibinfo{person}{Soroush Mehri}, \bibinfo{person}{Adriana
  Romero~Soriano}, \bibinfo{person}{Samira Shabanian}, {and}
  \bibinfo{person}{Sina Honari}.} \bibinfo{year}{2021}\natexlab{}.
\newblock \showarticletitle{Benchmarking Bias Mitigation Algorithms in
  Representation Learning through Fairness Metrics}. In
  \bibinfo{booktitle}{\emph{Proceedings of the Neural Information Processing
  Systems Track on Datasets and Benchmarks}},
  \bibfield{editor}{\bibinfo{person}{J.~Vanschoren} {and}
  \bibinfo{person}{S.~Yeung}} (Eds.), Vol.~\bibinfo{volume}{1}.
\newblock
\urldef\tempurl%
\url{https://datasets-benchmarks-proceedings.neurips.cc/paper/2021/file/2723d092b63885e0d7c260cc007e8b9d-Paper-round1.pdf}
\showURL{%
\tempurl}


\bibitem[Ritchie(2017)]%
        {Ritchie2017-RITSII}
\bibfield{author}{\bibinfo{person}{Katherine Ritchie}.}
  \bibinfo{year}{2017}\natexlab{}.
\newblock \showarticletitle{Social Identity, Indexicality, and the
  Appropriation of Slurs}.
\newblock \bibinfo{journal}{\emph{Croatian Journal of Philosophy}}
  \bibinfo{volume}{17}, \bibinfo{number}{2} (\bibinfo{year}{2017}),
  \bibinfo{pages}{155--180}.
\newblock


\bibitem[Romanov et~al\mbox{.}(2019)]%
        {romanov-what-name}
\bibfield{author}{\bibinfo{person}{Alexey Romanov}, \bibinfo{person}{Maria
  De-Arteaga}, \bibinfo{person}{Hanna Wallach}, \bibinfo{person}{Jennifer
  Chayes}, \bibinfo{person}{Christian Borgs}, \bibinfo{person}{Alexandra
  Chouldechova}, \bibinfo{person}{Sahin Geyik}, \bibinfo{person}{Krishnaram
  Kenthapadi}, \bibinfo{person}{Anna Rumshisky}, {and}
  \bibinfo{person}{Adam~Tauman Kalai}.} \bibinfo{year}{2019}\natexlab{}.
\newblock \bibinfo{title}{What's in a Name? Reducing Bias in Bios without
  Access to Protected Attributes}.
\newblock
\newblock
\urldef\tempurl%
\url{https://doi.org/10.48550/ARXIV.1904.05233}
\showDOI{\tempurl}


\bibitem[Ross et~al\mbox{.}(2021)]%
        {ross-etal-2021-measuring}
\bibfield{author}{\bibinfo{person}{Candace Ross}, \bibinfo{person}{Boris Katz},
  {and} \bibinfo{person}{Andrei Barbu}.} \bibinfo{year}{2021}\natexlab{}.
\newblock \showarticletitle{Measuring Social Biases in Grounded Vision and
  Language Embeddings}. In \bibinfo{booktitle}{\emph{Proceedings of the 2021
  Conference of the North American Chapter of the Association for Computational
  Linguistics: Human Language Technologies}}. \bibinfo{publisher}{Association
  for Computational Linguistics}, \bibinfo{address}{Online},
  \bibinfo{pages}{998--1008}.
\newblock
\urldef\tempurl%
\url{https://doi.org/10.18653/v1/2021.naacl-main.78}
\showDOI{\tempurl}


\bibitem[Ruder et~al\mbox{.}(2022)]%
        {ruder-etal-2022-square}
\bibfield{author}{\bibinfo{person}{Sebastian Ruder}, \bibinfo{person}{Ivan
  Vuli{\'c}}, {and} \bibinfo{person}{Anders S{\o}gaard}.}
  \bibinfo{year}{2022}\natexlab{}.
\newblock \showarticletitle{Square One Bias in {NLP}: Towards a
  Multi-Dimensional Exploration of the Research Manifold}. In
  \bibinfo{booktitle}{\emph{Findings of the Association for Computational
  Linguistics: ACL 2022}}. \bibinfo{publisher}{Association for Computational
  Linguistics}, \bibinfo{address}{Dublin, Ireland},
  \bibinfo{pages}{2340--2354}.
\newblock
\urldef\tempurl%
\url{https://doi.org/10.18653/v1/2022.findings-acl.184}
\showDOI{\tempurl}


\bibitem[Sap et~al\mbox{.}(2019)]%
        {sap-etal-2019-risk}
\bibfield{author}{\bibinfo{person}{Maarten Sap}, \bibinfo{person}{Dallas Card},
  \bibinfo{person}{Saadia Gabriel}, \bibinfo{person}{Yejin Choi}, {and}
  \bibinfo{person}{Noah~A. Smith}.} \bibinfo{year}{2019}\natexlab{}.
\newblock \showarticletitle{The Risk of Racial Bias in Hate Speech Detection}.
  In \bibinfo{booktitle}{\emph{Proceedings of the 57th Annual Meeting of the
  Association for Computational Linguistics}}. \bibinfo{publisher}{Association
  for Computational Linguistics}, \bibinfo{address}{Florence, Italy},
  \bibinfo{pages}{1668--1678}.
\newblock
\urldef\tempurl%
\url{https://doi.org/10.18653/v1/P19-1163}
\showDOI{\tempurl}


\bibitem[Shah et~al\mbox{.}(2020)]%
        {shah-etal-2020-predictive}
\bibfield{author}{\bibinfo{person}{Deven~Santosh Shah},
  \bibinfo{person}{H.~Andrew Schwartz}, {and} \bibinfo{person}{Dirk Hovy}.}
  \bibinfo{year}{2020}\natexlab{}.
\newblock \showarticletitle{Predictive Biases in Natural Language Processing
  Models: A Conceptual Framework and Overview}. In
  \bibinfo{booktitle}{\emph{Proceedings of the 58th Annual Meeting of the
  Association for Computational Linguistics}}. \bibinfo{publisher}{Association
  for Computational Linguistics}, \bibinfo{address}{Online},
  \bibinfo{pages}{5248--5264}.
\newblock
\urldef\tempurl%
\url{https://doi.org/10.18653/v1/2020.acl-main.468}
\showDOI{\tempurl}


\bibitem[Shen et~al\mbox{.}(2022)]%
        {shen-etal-2022-representational}
\bibfield{author}{\bibinfo{person}{Aili Shen}, \bibinfo{person}{Xudong Han},
  \bibinfo{person}{Trevor Cohn}, \bibinfo{person}{Timothy Baldwin}, {and}
  \bibinfo{person}{Lea Frermann}.} \bibinfo{year}{2022}\natexlab{}.
\newblock \showarticletitle{Does Representational Fairness Imply Empirical
  Fairness?}. In \bibinfo{booktitle}{\emph{Findings of the Association for
  Computational Linguistics: AACL-IJCNLP 2022}}.
  \bibinfo{publisher}{Association for Computational Linguistics},
  \bibinfo{address}{Online only}, \bibinfo{pages}{81--95}.
\newblock
\urldef\tempurl%
\url{https://aclanthology.org/2022.findings-aacl.8}
\showURL{%
\tempurl}


\bibitem[Stanczak and Augenstein(2021)]%
        {karolina-survey}
\bibfield{author}{\bibinfo{person}{Karolina Stanczak} {and}
  \bibinfo{person}{Isabelle Augenstein}.} \bibinfo{year}{2021}\natexlab{}.
\newblock \bibinfo{title}{A Survey on Gender Bias in Natural Language
  Processing}.
\newblock
\newblock
\urldef\tempurl%
\url{https://doi.org/10.48550/ARXIV.2112.14168}
\showDOI{\tempurl}


\bibitem[Sun et~al\mbox{.}(2019)]%
        {sun-etal-2019-mitigating}
\bibfield{author}{\bibinfo{person}{Tony Sun}, \bibinfo{person}{Andrew Gaut},
  \bibinfo{person}{Shirlyn Tang}, \bibinfo{person}{Yuxin Huang},
  \bibinfo{person}{Mai ElSherief}, \bibinfo{person}{Jieyu Zhao},
  \bibinfo{person}{Diba Mirza}, \bibinfo{person}{Elizabeth Belding},
  \bibinfo{person}{Kai-Wei Chang}, {and} \bibinfo{person}{William~Yang Wang}.}
  \bibinfo{year}{2019}\natexlab{}.
\newblock \showarticletitle{Mitigating Gender Bias in Natural Language
  Processing: Literature Review}. In \bibinfo{booktitle}{\emph{Proceedings of
  the 57th Annual Meeting of the Association for Computational Linguistics}}.
  \bibinfo{publisher}{Association for Computational Linguistics},
  \bibinfo{address}{Florence, Italy}, \bibinfo{pages}{1630--1640}.
\newblock
\urldef\tempurl%
\url{https://doi.org/10.18653/v1/P19-1159}
\showDOI{\tempurl}


\bibitem[Tan and Celis(2019)]%
        {tan-and-celis}
\bibfield{author}{\bibinfo{person}{Yi~Chern Tan} {and}
  \bibinfo{person}{L.~Elisa Celis}.} \bibinfo{year}{2019}\natexlab{}.
\newblock \bibinfo{booktitle}{\emph{Assessing Social and Intersectional Biases
  in Contextualized Word Representations}}.
\newblock \bibinfo{publisher}{Curran Associates Inc.}, \bibinfo{address}{Red
  Hook, NY, USA}.
\newblock


\bibitem[Vulić et~al\mbox{.}(2020)]%
        {vulic-multisimlex}
\bibfield{author}{\bibinfo{person}{Ivan Vulić}, \bibinfo{person}{Simon Baker},
  \bibinfo{person}{Edoardo~Maria Ponti}, \bibinfo{person}{Ulla Petti},
  \bibinfo{person}{Ira Leviant}, \bibinfo{person}{Kelly Wing},
  \bibinfo{person}{Olga Majewska}, \bibinfo{person}{Eden Bar},
  \bibinfo{person}{Matt Malone}, \bibinfo{person}{Thierry Poibeau},
  \bibinfo{person}{Roi Reichart}, {and} \bibinfo{person}{Anna Korhonen}.}
  \bibinfo{year}{2020}\natexlab{}.
\newblock \showarticletitle{{Multi-SimLex: A Large-Scale Evaluation of
  Multilingual and Crosslingual Lexical Semantic Similarity}}.
\newblock \bibinfo{journal}{\emph{Computational Linguistics}}
  \bibinfo{volume}{46}, \bibinfo{number}{4} (\bibinfo{date}{02}
  \bibinfo{year}{2020}), \bibinfo{pages}{847--897}.
\newblock
\showISSN{0891-2017}
\urldef\tempurl%
\url{https://doi.org/10.1162/coli_a_00391}
\showDOI{\tempurl}
\showeprint{https://direct.mit.edu/coli/article-pdf/46/4/847/1888287/coli\_a\_00391.pdf}


\bibitem[Wei and Santos~Jr.(2020)]%
        {Wei_SantosJr2020}
\bibfield{author}{\bibinfo{person}{Jason Wei} {and} \bibinfo{person}{Eugene
  Santos~Jr.}} \bibinfo{year}{2020}\natexlab{}.
\newblock \showarticletitle{Narrative Origin Classification of
  {I}sraeli-{P}alestinian Conflict Texts}. In \bibinfo{booktitle}{\emph{The
  Thirty-Third International FLAIRS Conference}}.
\newblock


\bibitem[Williamson and Menon(2019)]%
        {pmlr-v97-williamson19a}
\bibfield{author}{\bibinfo{person}{Robert Williamson} {and}
  \bibinfo{person}{Aditya Menon}.} \bibinfo{year}{2019}\natexlab{}.
\newblock \showarticletitle{Fairness risk measures}. In
  \bibinfo{booktitle}{\emph{Proceedings of the 36th International Conference on
  Machine Learning}} \emph{(\bibinfo{series}{Proceedings of Machine Learning
  Research}, Vol.~\bibinfo{volume}{97})},
  \bibfield{editor}{\bibinfo{person}{Kamalika Chaudhuri} {and}
  \bibinfo{person}{Ruslan Salakhutdinov}} (Eds.). \bibinfo{publisher}{PMLR},
  \bibinfo{pages}{6786--6797}.
\newblock
\urldef\tempurl%
\url{https://proceedings.mlr.press/v97/williamson19a.html}
\showURL{%
\tempurl}


\bibitem[Zhang et~al\mbox{.}(2021)]%
        {zhang-etal-2021-sociolectal}
\bibfield{author}{\bibinfo{person}{Sheng Zhang}, \bibinfo{person}{Xin Zhang},
  \bibinfo{person}{Weiming Zhang}, {and} \bibinfo{person}{Anders S{\o}gaard}.}
  \bibinfo{year}{2021}\natexlab{}.
\newblock \showarticletitle{Sociolectal Analysis of Pretrained Language
  Models}. In \bibinfo{booktitle}{\emph{Proceedings of the 2021 Conference on
  Empirical Methods in Natural Language Processing}}.
  \bibinfo{publisher}{Association for Computational Linguistics},
  \bibinfo{address}{Online and Punta Cana, Dominican Republic},
  \bibinfo{pages}{4581--4588}.
\newblock
\urldef\tempurl%
\url{https://doi.org/10.18653/v1/2021.emnlp-main.375}
\showDOI{\tempurl}


\bibitem[Zhao et~al\mbox{.}(2018)]%
        {zhao-etal-2018-learning}
\bibfield{author}{\bibinfo{person}{Jieyu Zhao}, \bibinfo{person}{Yichao Zhou},
  \bibinfo{person}{Zeyu Li}, \bibinfo{person}{Wei Wang}, {and}
  \bibinfo{person}{Kai-Wei Chang}.} \bibinfo{year}{2018}\natexlab{}.
\newblock \showarticletitle{Learning Gender-Neutral Word Embeddings}. In
  \bibinfo{booktitle}{\emph{Proceedings of the 2018 Conference on Empirical
  Methods in Natural Language Processing}}. \bibinfo{publisher}{Association for
  Computational Linguistics}, \bibinfo{address}{Brussels, Belgium},
  \bibinfo{pages}{4847--4853}.
\newblock
\urldef\tempurl%
\url{https://doi.org/10.18653/v1/D18-1521}
\showDOI{\tempurl}


\end{thebibliography}

\appendix

\end{document}